\documentclass[a4paper,fleqn]{cas-sc}

\usepackage[authoryear,longnamesfirst]{natbib}

\usepackage{subfig}
\usepackage{siunitx}
\usepackage{booktabs}
\usepackage{tabularx}
\usepackage{array}
\usepackage{makecell}
\usepackage{capt-of}

\def\tsc#1{\csdef{#1}{\textsc{\lowercase{#1}}\xspace}}
\tsc{WGM}
\tsc{QE}

\begin{document}
\let\WriteBookmarks\relax
\def\floatpagepagefraction{1}
\def\textpagefraction{.001}
\let\printorcid\relax

\shorttitle{}    

\shortauthors{}  

\title [mode = title]{A Compact Hybrid Convolution--Frequency State Space Network for Learned Image Compression}  



%

\author[1]{Haodong Pan}



\ead{hd.pan@stu.xjtu.edu.cn}


\credit{Writing - review \& editing, Writing - original draft, Visualization, Validation, Conceptualization, Methodology, Investigation, Formal analysis, Software, Data curation}

\affiliation[1]{organization={State Key Laboratory of Human-Machine Hybrid Augmented Intelligence, Institute of Artificial Intelligence and Robotics, Xi'an Jiaotong University},
            city={Xi'an},
            postcode={710049}, 
            state={Shaanxi},
            country={China}}

\author[1]{Hao Wei}


\ead{haowei@stu.xjtu.edu.cn}


\credit{Writing - Review \& Editing, Conceptualization, Visualization, Validation, Formal analysis, Data curation}


\author[1]{Yusong Wang}

\ead{wangyusong2000@stu.xjtu.edu.cn}
\credit{Writing - Review \& Editing, Conceptualization}

\author[1]{Nanning Zheng}


\ead{nnzheng@mail.xjtu.edu.cn}


\credit{Writing - Review \& Editing, Supervision, Resources, Funding acquisition}

\author[1]{Caigui Jiang}

\cormark[1]


\ead{cgjiang@xjtu.edu.cn}


\credit{Writing - Review \& Editing, Supervision, Project administration, Funding acquisition}

\cortext[1]{Corresponding author}




\begin{abstract}
Learned image compression (LIC) has recently benefited from Transformer- and state space models (SSM)- based backbones for modeling long-range dependencies. However, the former typically incurs quadratic complexity, whereas the latter often disrupts neighborhood continuity by flattening 2D features into 1D sequences. To address these issues, we propose a compact Hybrid Convolution and Frequency State Space Network (HCFSSNet) for LIC. HCFSSNet combines convolutional layers for local detail modeling with a Vision Frequency State Space (VFSS) block for complementary long-range contextual aggregation. Specifically, the VFSS block consists of a Vision Omni-directional Neighborhood State Space (VONSS) module, which scans features along horizontal, vertical, and diagonal directions to better preserve 2D neighborhood relations, and an Adaptive Frequency Modulation Module (AFMM), which performs discrete cosine transform-based adaptive reweighting of frequency components. In addition, we introduce a Frequency Swin Transformer Attention Module (FSTAM) in the hyperprior path to enhance frequency-aware side information modeling. Experiments on the benchmark datasets show that the proposed HCFSSNet achieves a competitive rate-distortion performance against recent LIC codecs. The source code and models will be made publicly available.
\end{abstract}





\maketitle

\section{Introduction}

Image compression is a fundamental problem in computer vision and multimedia applications, as it directly affects storage efficiency, transmission cost, and downstream visual analysis. Over the past decades, traditional image compression techniques have led to widely adopted standards such as JPEG~\cite{wallace1992jpeg}, JPEG2000~\cite{skodras2001jpeg2000}, WebP~\cite{ginesu2012objective}, BPG~\cite{bellard2014bpg}, and VVC~\cite{bross2021overview}. These hand-designed codecs rely on carefully engineered transforms, quantization strategies, and entropy coding pipelines, but they are less flexible in modeling the complex statistical dependencies of natural images. With the rapid progress of deep learning, learned image compression (LIC) has attracted increasing attention. By jointly optimizing the analysis transform, synthesis transform, and entropy model in an end-to-end manner, LIC methods~\cite{xie2021enhanced, liu2023learned, li2024frequencyaware} have achieved strong rate--distortion performance and become an important research direction in modern image compression.

Most LIC methods are built on an autoencoder framework~\cite{Kingma2014autoencoder}, which typically consists of three key components: a nonlinear transform, quantization, and an entropy model. During encoding, the input image is mapped by an analysis transform to a compact latent representation. The latent is then discretized by quantization, and an entropy model estimates its probability distribution for subsequent lossless coding, such as arithmetic coding~\cite{rissanen1981universal} or range coding~\cite{martin1979range}. During decoding, the bitstream is entropy-decoded to recover the discrete latent representation, which is then mapped back to the reconstructed image by the synthesis transform.

\begin{center}
\begin{minipage}{\linewidth}
\centering
\includegraphics[width=0.6\linewidth]{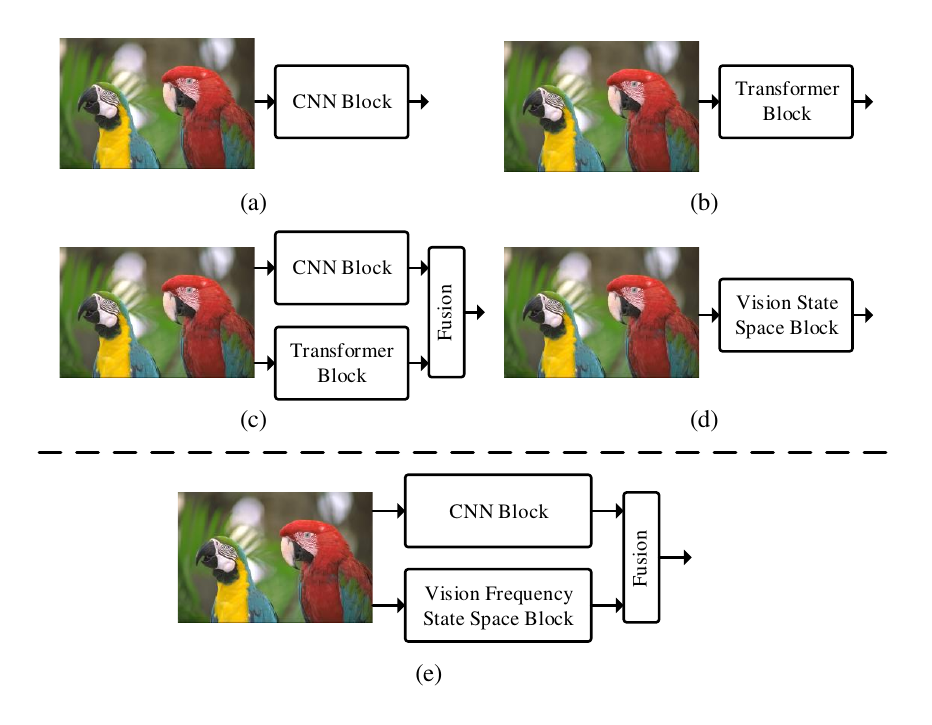}
\captionof{figure}{Representative nonlinear transforms used in learned image compression. (a) CNN-based methods (e.g., Chen et al.~\cite{chen2021nonlocal}); (b) Transformer-based methods (e.g., Zhu et al.~\cite{zhu2022transformerbased}); (c) hybrid CNN--Transformer methods (e.g., Liu et al.~\cite{liu2023learned}); (d) vision state space model (SSM)-based methods (e.g., Qin et al.~\cite{qin2024mambavc}); and (e) the proposed HCFSSNet, which combines a CNN branch for local detail modeling with a Vision Frequency State Space (VFSS) branch for complementary long-range contextual modeling.}
\label{fig:0}
\end{minipage}
\end{center}

Recent progress in LIC has largely been driven by the design of stronger nonlinear transforms. As illustrated in Fig.~\ref{fig:0}, existing approaches can be broadly grouped into four categories: (a) CNN-based methods, (b) Transformer-based methods, (c) hybrid CNN--Transformer methods, and (d) vision state space model (SSM)-based methods. CNN-based methods, such as the non-local attention network of Chen et al.~\cite{chen2021nonlocal}, are effective at capturing local structures while incorporating limited long-range interactions. Transformer-based methods, represented by the Swin-Transformer-based framework of Zhu et al.~\cite{zhu2022transformerbased}, introduce window-based self-attention to model broader contextual dependencies. Hybrid CNN--Transformer methods further combine the strengths of local convolution and global attention, as in the mixed-block design of Liu et al.~\cite{liu2023learned}. More recently, vision SSM-based methods, such as MambaVC~\cite{qin2024mambavc} and MambaIC~\cite{zeng2025mambaic}, have shown that state space models can provide effective long-range modeling with linear complexity.

Despite these advances, several challenges remain when adapting these architectures to LIC. Transformer-based models are effective in capturing non-local dependencies, but their quadratic complexity with respect to sequence length limits scalability on high-resolution images. Vision SSMs offer a more efficient alternative, but extending sequence models from 1D signals to 2D feature maps is non-trivial. In particular, directly flattening 2D features into 1D sequences may weaken neighborhood continuity and make some spatial relations harder to preserve. Existing vision SSM variants such as Vim~\cite{Zhu2024vim}, VMamba~\cite{liu2024vmamba}, and MambaIC~\cite{zeng2025mambaic} mainly rely on horizontal and vertical scan patterns. As a result, some 2D neighborhood interactions, especially along diagonal directions, may be less explicitly modeled, which is not ideal for compression-oriented representation learning.

In parallel with advances in nonlinear transforms, another important research line has focused on entropy modeling for more accurate bitrate estimation. Hyperprior-based approaches~\cite{minnen2018joint, zhou2019multiscale, cui2021asymmetric, li2024frequencyaware} employ masked convolutions~\cite{oord2016conditional} or masked Transformers~\cite{devlin2019bert} to model dependencies in quantized latents, while 3D masked convolutions have been explored to jointly capture spatial and cross-channel dependencies~\cite{qi2016volumetric, ji20123d, Chen2021end, mentzer2018conditional, tang2023joint}. In addition, channel-wise autoregressive schemes~\cite{minnen2020channel} and spatial--channel context-adaptive models~\cite{he2022elic} have been proposed to further reduce redundancy. Some recent studies have also introduced frequency-aware components into LIC models~\cite{li2024frequencyaware, wang2024fdnet}. Nevertheless, most entropy models still operate primarily in the spatial and channel domains, and frequency-aware modeling of side information remains relatively underexplored. A finer-grained treatment of different frequency components within latent and hyperprior features may therefore provide complementary benefits for LIC.

Overall, current LIC frameworks have substantially improved rate--distortion performance, but three issues remain worth further investigation. First, existing visual SSMs are not fully tailored to 2D neighborhood relations and may insufficiently model diagonal interactions. Second, many LIC architectures do not jointly balance local detail modeling and long-range contextual aggregation within a compact unified design. Third, hyperprior-based entropy models rarely incorporate fine-grained frequency-aware modeling of side information.

To address these issues, we present a compact Hybrid Convolution and Frequency State Space Network (HCFSSNet) for learned image compression. As shown in Fig.~\ref{fig:0}(e), HCFSSNet combines convolution-based local detail modeling with state-space-based long-range contextual modeling in a unified framework. In the main transform, we design a Hybrid Convolution--Frequency State Space (HCFSS) block that splits feature channels into two branches: a CNN branch for local representation refinement and a Vision Frequency State Space (VFSS) branch for complementary contextual aggregation. To better adapt SSMs to 2D image features, the VFSS branch incorporates a Vision Omni-directional Neighborhood State Space (VONSS) module, which scans features along horizontal, vertical, diagonal, and anti-diagonal directions to better preserve 2D neighborhood relations. In addition, we introduce an Adaptive Frequency Modulation Module (AFMM), which performs discrete-cosine-transform-based adaptive reweighting of frequency components. For entropy modeling, AFMM is further integrated with a Swin Transformer block to form a Frequency Swin Transformer Attention Module (FSTAM), which enhances frequency-aware side-information modeling in the hyperprior path.

Very recently, LIC models such as MambaIC~\cite{zeng2025mambaic}, DCAE~\cite{lu2025learned}, and LALIC~\cite{feng2025linear} have continued to improve rate--distortion performance through increasingly sophisticated sequence and context modeling modules. These methods demonstrate the strong potential of high-capacity architectures, but they also move toward more complex design choices. In contrast, our goal is not to push BD-rate to the absolute minimum with the heaviest possible architecture. Instead, we aim to develop a compact hybrid design that integrates local spatial modeling, long-range dependency modeling, and frequency-aware modulation within a unified LIC framework.

The main contributions of this work are summarized as follows:
\begin{itemize}
    \item We propose HCFSSNet, a compact hybrid convolution--state-space architecture for learned image compression, which combines local spatial detail modeling with long-range contextual modeling in a unified framework.

    \item We design a Vision Frequency State Space (VFSS) block that integrates an omni-directional neighborhood scanning strategy (VONSS) with a DCT-based Adaptive Frequency Modulation Module (AFMM) for compression-oriented representation learning.

    \item We introduce a frequency-aware hyperprior module, termed FSTAM, and show through experiments and ablations on Kodak, Tecnick, and CLIC Professional Validation that HCFSSNet achieves competitive rate--distortion performance with fewer parameters than recent SSM- and Transformer-based LIC codecs.
\end{itemize}

\section{Related Work}

\subsection{Learned Image Compression}

Learned image compression (LIC) has become a major research direction in image compression by jointly optimizing the nonlinear transform, quantization surrogate, and entropy model in an end-to-end manner. Existing LIC methods can be roughly grouped according to two key aspects: the design of the nonlinear transform and the design of the entropy model.

\subsubsection{Nonlinear Transform}

Early LIC methods mainly relied on convolutional neural networks (CNNs) to construct nonlinear analysis and synthesis transforms. Ballé et al.~\cite{ballé2017endtoend} introduced one of the earliest end-to-end CNN-based compression frameworks with generalized divisive normalization. Subsequent works enhanced convolutional backbones with more expressive context modeling modules. For example, Chen et al.~\cite{Chen2021end} incorporated non-local attention to improve the representation of informative structures, while Tang et al.~\cite{tang2023joint} combined asymmetric convolutions with graph attention to capture both local and non-local dependencies. These studies show that CNN-based transforms provide a strong foundation for LIC, especially for local structure modeling. In addition, recent analyses have investigated the stability and optimization behavior of learned image compression models under successive training schemes, highlighting the importance of architectural and training design in achieving reliable performance~\cite{kim2022successive}.

Transformer-based LIC methods further improved contextual modeling by introducing self-attention. Zhu et al.~\cite{zhu2022transformerbased} developed a Swin-Transformer-based framework for image and video compression, where window-based self-attention enables broader dependency modeling. Several subsequent methods combined Transformers with convolutions to exploit both local inductive bias and long-range context~\cite{zou2022devil, lu2022transformer, liu2023learned}. More recently, some studies have also explored frequency-aware designs within this family. For instance, Li et al.~\cite{li2024frequencyaware} introduced frequency-aware attention and modulation modules, and Wang et al.~\cite{wang2024frequency} proposed a hybrid architecture that processes different frequency components with different branches. These methods demonstrate the benefit of combining stronger context modeling with frequency-sensitive representations.

More recently, state space models (SSMs) have emerged as an efficient alternative for long-range visual modeling. Motivated by the linear-time sequence modeling capability of Mamba~\cite{gu2024mamba}, Qin et al.~\cite{qin2024mambavc} introduced Mamba-based modules into image and video compression, and Zeng et al.~\cite{zeng2025mambaic} further developed an SSM-based LIC framework that integrates visual state space blocks into both the main transform and entropy-related modules. In parallel, vision SSM variants such as Vim~\cite{Zhu2024vim} and VMamba~\cite{liu2024vmamba} extended 1D SSMs to 2D visual features through directional scan strategies. These works suggest that SSMs are promising for efficient long-range dependency modeling in LIC. However, existing visual SSM designs are still largely adapted from sequence modeling and typically rely on horizontal and vertical scan patterns, while the modeling of full 2D neighborhood relations and frequency-aware representations remains limited.

\subsubsection{Entropy Model}

Another central component of LIC is the entropy model, which estimates the probability distribution of quantized latent representations for bitrate coding. A major advance in this direction was the hyperprior framework of Ballé et al.~\cite{ballé2018variational}, which introduced side information to better capture spatially varying latent statistics. Minnen et al.~\cite{minnen2018joint} further showed that autoregressive and hierarchical priors are complementary, establishing a widely adopted hyperprior-plus-context paradigm in LIC.

To improve context modeling, a large body of work has explored different forms of spatial and channel dependency modeling. Checkerboard context models~\cite{he2021checkerboard} were proposed to alleviate the sequential bottleneck of fully autoregressive decoding. Mentzer et al.~\cite{mentzer2018conditional} explored 3D CNN-based entropy models for cross-channel dependency modeling, while Minnen et al.~\cite{minnen2020channel} introduced channel-wise autoregressive priors based on channel slices. Building on these ideas, He et al.~\cite{he2022elic} proposed an unevenly grouped spatial--channel entropy model, and Li et al.~\cite{li2024groupedmixer} further improved joint spatial--channel modeling through grouped context modeling. More recently, entropy models with stronger context aggregation have continued to improve probability estimation accuracy~\cite{wang2025s2lic}, while subsequent studies further explored more expressive entropy modeling strategies, including multi-reference entropy models and learned prior generation mechanisms that improve probability estimation through richer contextual dependencies~\cite{he2026depth, zhao2025deeply}.

Overall, existing entropy models have substantially improved bitrate estimation by exploiting spatial, channel, and hyperprior information. Nevertheless, most of these designs still operate primarily in the spatial and channel domains, while frequency-aware modeling of side information in the hyperprior pathway has received relatively limited attention.

\subsection{Frequency-aware Modeling for Image Compression}

Frequency-domain analysis has long played an important role in traditional image compression. Classical codecs such as JPEG, JPEG2000, WebP, and BPG rely on transforms such as the discrete cosine transform (DCT) or discrete wavelet transform (DWT) to convert images into transform domains where different frequency components can be processed differently. This strategy provides an effective inductive bias for compression because image content often exhibits different statistics across frequency bands.

Inspired by this observation, recent learned compression methods have begun to incorporate frequency-aware modeling. Li et al.~\cite{li2024frequencyaware} proposed frequency-aware attention and modulation modules to better capture anisotropic frequency information in LIC. Wang et al.~\cite{wang2024fdnet} introduced dynamic frequency filtering to separate feature components and process them with different modules. 

Beyond compression, frequency-domain modeling has also been explored in broader vision tasks. For example, Wave-ViT~\cite{yao2022wave} integrates wavelet transforms with transformer architectures to enhance multi-scale representation learning, while Tang et al.~\cite{tang2025spatial} combine spatial and frequency-domain processing for improved image restoration. These studies suggest that transform-domain representations can provide complementary benefits to purely spatial modeling.

Complementary efforts in learned compression also focus on improving prior modeling and entropy estimation, further emphasizing the importance of combining representation learning with accurate probability modeling in learned compression frameworks~\cite{he2026depth, zhao2025deeply}.

However, existing frequency-aware LIC methods are still limited in two aspects. First, many methods perform relatively coarse frequency decomposition or use frequency modules as auxiliary components, rather than integrating them tightly with the main long-range modeling pathway. Second, frequency-aware modeling is rarely extended to the hyperprior branch for side-information refinement. In contrast, our work is motivated by incorporating frequency-aware modulation into both the main transform and the hyperprior pathway within a unified hybrid architecture.

Compared with recent visual SSM-based methods such as Vim and VMamba, which extend sequence-based scanning strategies to 2D data, our design explicitly targets the mismatch between sequential scan order and spatial neighborhood relations. In particular, the proposed omni-directional scan improves the preservation of 2D local structures, including diagonal dependencies that are often weakened under conventional horizontal or vertical scan patterns. Compared with recent frequency-aware LIC approaches such as FTIC, which mainly introduce frequency modeling within the main transform, our method integrates frequency-aware modulation more tightly into both the main transform and the hyperprior pathway. This allows frequency-selective refinement not only for feature representation but also for side-information modeling in entropy estimation. Finally, compared with high-capacity models such as MambaIC, DCAE, and related large-scale architectures that aim to push BD-rate to the extreme, our goal is to achieve a more compact and unified design. HCFSSNet balances model size, architectural simplicity, and competitive rate--distortion performance, rather than optimizing a single metric in isolation.

\section{Method}

\subsection{State Space Model Preliminaries}

State space models (SSMs) are commonly formulated as linear time-invariant (LTI) systems that map a one-dimensional input signal \(x(t) \in \mathbb{R}\) to an output signal \(y(t) \in \mathbb{R}\) through an \(n\)-dimensional hidden state \(\mathsf{h}(t) \in \mathbb{R}^n\). The continuous-time dynamics can be written as
\begin{equation}
\begin{aligned}
    \mathsf{h}'(t) &= \mathbf{A}\mathsf{h}(t) + \mathbf{B}x(t), \\
    y(t) &= \mathbf{C}\mathsf{h}(t) + \mathbf{D}x(t),
\end{aligned}
\label{eq:eq1}
\end{equation}
where \(\mathbf{A} \in \mathbb{R}^{n \times n}\) is the state matrix, \(\mathbf{B} \in \mathbb{R}^{n \times 1}\) is the input matrix, \(\mathbf{C} \in \mathbb{R}^{1 \times n}\) is the output matrix, and \(\mathbf{D} \in \mathbb{R}\) is the feed-through term. These parameters are learnable and can be optimized end-to-end.

For practical deep learning applications, the continuous system is discretized using the zero-order hold (ZOH) rule. The matrices \(\mathbf{A}\) and \(\mathbf{B}\) are converted into their discrete counterparts as
\begin{equation}
\begin{aligned}
\overline{\mathbf{A}} &= \exp(\Delta \mathbf{A}), \\
\overline{\mathbf{B}} &= (\Delta \mathbf{A})^{-1}\left(\exp(\Delta \mathbf{A}) - \mathbf{I}\right)\Delta \mathbf{B},
\end{aligned}
\end{equation}
where \(\Delta\) denotes the discretization step, \(\exp(\cdot)\) is the matrix exponential, and \(\mathbf{I}\) is the identity matrix. The resulting discrete-time formulation is
\begin{equation}
\begin{aligned}
\mathsf{h}_t &= \overline{\mathbf{A}} \mathsf{h}_{t-1} + \overline{\mathbf{B}} x_t, \\
y_t &= \mathbf{C}\mathsf{h}_t + \mathbf{D}x_t.
\end{aligned}
\label{eq:eq3}
\end{equation}

Recent selective SSMs further improve the practicality of this formulation in deep neural networks. In particular, Mamba~\cite{gu2024mamba} introduces input-dependent selective scanning together with an efficient parallel implementation based on the discrete-time formulation in Eq.~\ref{eq:eq3}. This design provides effective long-range sequence modeling with linear complexity in the sequence length. In this work, we build our visual state space design on this family of modern selective SSMs and adapt it to two-dimensional image features.

\subsection{Problem Formulation}

\begin{center}
\begin{minipage}{\textwidth}
\centering
\includegraphics[width=0.85\linewidth]{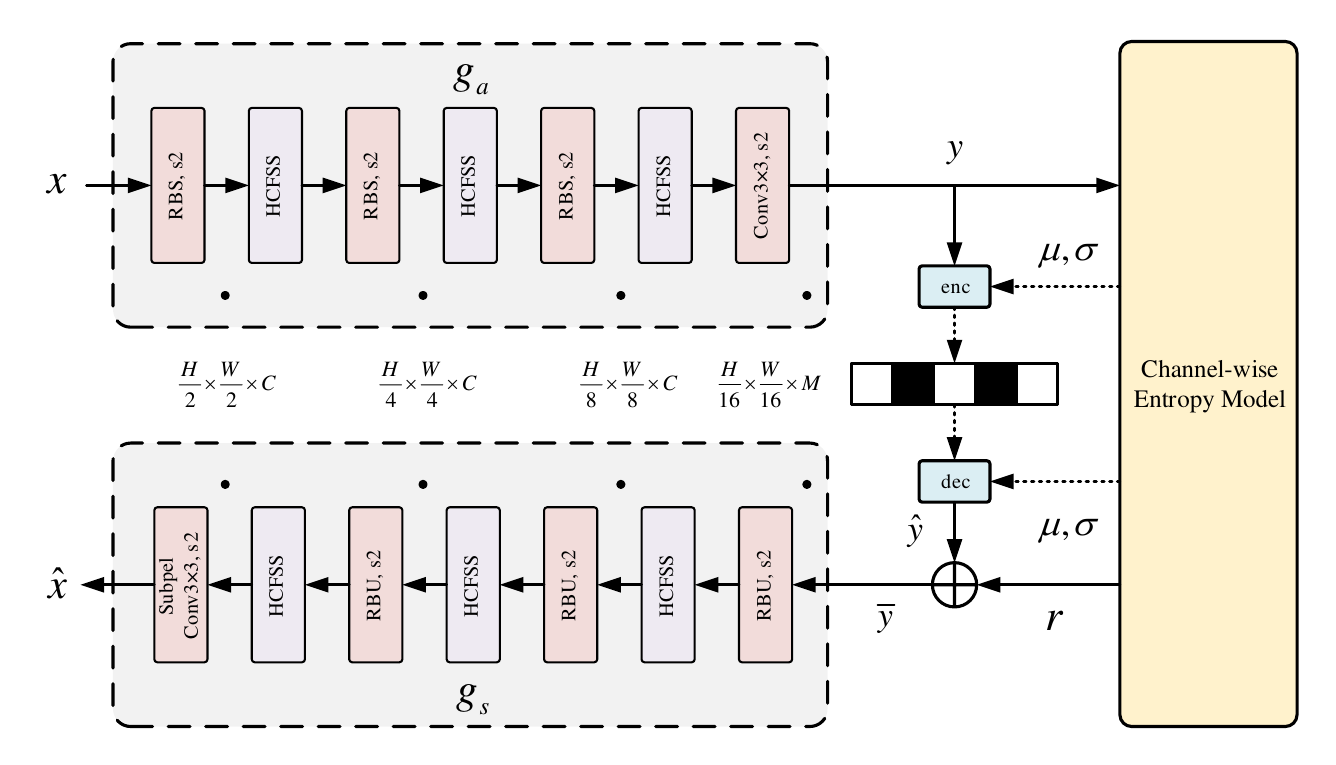}
\captionof{figure}{Overall architecture of the proposed HCFSSNet. The HCFSS block denotes the Hybrid Convolution--Frequency State Space block. RBS and RBU denote residual blocks for downsampling and upsampling, respectively, built from \(1\times1\) and \(3\times3\) convolutions. ``s2'' denotes a stride of 2, while ``enc'' and ``dec'' denote range encoders/decoders with quantization. \(C\) and \(M\) denote the numbers of channels used for feature extraction and latent representation, respectively.}
\label{fig:1}
\end{minipage}
\end{center}

Following standard hyperprior-based learned image compression (LIC) frameworks, the proposed HCFSSNet consists of an analysis transform, a synthesis transform, a hyperprior pathway, and a channel-wise entropy model, as shown in Fig.~\ref{fig:1}. Our main architectural modifications lie in the Hybrid Convolution--Frequency State Space (HCFSS) block used in the main and hyperprior transforms, and in the frequency-aware refinement module used in the entropy model.

Formally, given an input image \(\boldsymbol{x}\), the analysis transform \(g_a\) maps it to the latent representation \(\boldsymbol{y}\), and the hyper-analysis transform \(h_a\) produces the corresponding hyperprior \(\boldsymbol{z}\):
\begin{equation}
\begin{aligned}
    \boldsymbol{y} &= g_a(\boldsymbol{x}; \phi_g), \\
    \boldsymbol{z} &= h_a(\boldsymbol{y}; \phi_h), \\
    \hat{\boldsymbol{z}} &= Q(\boldsymbol{z} - \boldsymbol{m}) + \boldsymbol{m}, \\
    F_{\text{mean}} &= h_{\text{mean}}(\hat{\boldsymbol{z}}; \theta_{\text{mean}}), \\
    F_{\text{scale}} &= h_{\text{scale}}(\hat{\boldsymbol{z}}; \theta_{\text{scale}}), \\
    \bar{\boldsymbol{y}}_{i}, \mu_i, \sigma_i &= e_i(F_{\text{mean}}, F_{\text{scale}}, \bar{\boldsymbol{y}}_{<i}, \boldsymbol{y}_i), \quad 0 \le i < n, \\
    \hat{\boldsymbol{x}} &= g_s(\bar{\boldsymbol{y}}; \theta_s),
\end{aligned}
\end{equation}
where \(g_a\) and \(g_s\) denote the analysis and synthesis transforms parameterized by \(\phi_g\) and \(\theta_s\), respectively; \(h_a\) denotes the hyper-analysis transform parameterized by \(\phi_h\); \(h_{\text{mean}}\) and \(h_{\text{scale}}\) are two hyper-synthesis transforms parameterized by \(\theta_{\text{mean}}\) and \(\theta_{\text{scale}}\), respectively; and \(Q(\cdot)\) denotes a uniform scalar quantizer. The hyperprior \(\boldsymbol{z}\) is centered by subtracting its median \(\boldsymbol{m}\) and then quantized to \(\hat{\boldsymbol{z}}\).

Throughout the rest of this paper, bold lowercase letters (e.g., \(\boldsymbol{x}\)) denote vectors or tensors, while plain letters (e.g., \(x_{ij}\), \(\mu_i\), and \(\sigma_i\)) denote scalars or individual elements.

Following previous channel-slice entropy models~\cite{minnen2020channel, liu2023learned}, the latent representation \(\boldsymbol{y}\) is divided into \(n\) equally sized channel slices \(\{\boldsymbol{y}_0, \boldsymbol{y}_1, \ldots, \boldsymbol{y}_{n-1}\}\). For the \(i\)-th slice, the entropy model \(e_i\) takes as input the hyperprior features \(F_{\text{mean}}\) and \(F_{\text{scale}}\), the previously reconstructed slices \(\bar{\boldsymbol{y}}_{<i} = \{\bar{\boldsymbol{y}}_0, \ldots, \bar{\boldsymbol{y}}_{i-1}\}\), and the current slice \(\boldsymbol{y}_i\), and predicts its Gaussian parameters \(\mu_i\) and \(\sigma_i\) together with the refined reconstruction \(\bar{\boldsymbol{y}}_i\). The synthesis transform \(g_s\) then reconstructs the output image \(\hat{\boldsymbol{x}}\) from \(\bar{\boldsymbol{y}} = \{\bar{\boldsymbol{y}}_i\}_{i=0}^{n-1}\).

During encoding, the difference between the current slice \(\boldsymbol{y}_i\) and its predicted mean \(\mu_i\) is quantized as
\begin{equation}
\begin{aligned}
    \hat{\boldsymbol{y}}_{i} &= Q(\boldsymbol{y}_i - \mu_i) + \mu_i, \\
    \boldsymbol{r}_i &= \text{Res}_i(\mu_i, \hat{\boldsymbol{y}}_i), \\
    \bar{\boldsymbol{y}}_{i} &= \hat{\boldsymbol{y}}_{i} + \boldsymbol{r}_i,
\end{aligned}
\end{equation}
where \(\text{Res}_i(\cdot)\) denotes the residual prediction network for the \(i\)-th slice. It estimates the residual \(\boldsymbol{r}_i\) between \(\boldsymbol{y}_i\) and \(\hat{\boldsymbol{y}}_i\), and the refined slice \(\bar{\boldsymbol{y}}_i\) is obtained by adding this residual to \(\hat{\boldsymbol{y}}_i\).

Following Ballé et al.~\cite{Ballé2016density}, we use a factorized density model for the quantized hyperprior \(\hat{\boldsymbol{z}}\):
\begin{equation}
\begin{aligned}
p_{\hat{\boldsymbol{z}}|\boldsymbol{\psi}}(\hat{\boldsymbol{z}} \mid \boldsymbol{\psi})
= \prod_j \left( p_{z_j|\psi_j}(\psi_j) * \mathcal{U}\!\left(-\frac{1}{2}, \frac{1}{2}\right) \right)(\hat{z}_j),
\end{aligned}
\end{equation}
where \(\boldsymbol{\psi} = \{\psi_j\}\) denotes the hyperprior parameters and \(\mathcal{U}(-\tfrac{1}{2}, \tfrac{1}{2})\) models the quantization noise. Similarly, following~\cite{ballé2018variational}, each quantized latent coefficient \(\hat{y}_j\) is modeled by a Gaussian distribution with mean \(\mu_j\) and standard deviation \(\sigma_j\):
\begin{equation}
\begin{aligned}
p_{\hat{\boldsymbol{y}}|\hat{\boldsymbol{z}}}(\hat{\boldsymbol{y}} \mid \hat{\boldsymbol{z}}, \theta)
= \prod_j \left( \mathcal{N}(\mu_j, \sigma_j^2) * \mathcal{U}\!\left(-\frac{1}{2}, \frac{1}{2}\right) \right)(\hat{y}_j),
\end{aligned}
\end{equation}
where \(\hat{\boldsymbol{y}} = \{\hat{y}_j\}\) collects all quantized latent coefficients. The integer-valued outputs of the quantizers, \(Q(\boldsymbol{z} - \boldsymbol{m})\) and \(Q(\boldsymbol{y} - \boldsymbol{\mu})\), are entropy-coded into a bitstream using lossless coding methods such as arithmetic coding. The reconstructed latents are then given by \(\hat{\boldsymbol{z}} = Q(\boldsymbol{z} - \boldsymbol{m}) + \boldsymbol{m}\) and \(\hat{\boldsymbol{y}} = Q(\boldsymbol{y} - \boldsymbol{\mu}) + \boldsymbol{\mu}\).

To optimize the rate--distortion trade-off, we adopt the following Lagrangian objective:
\begin{equation}
\begin{split}
L &= R(\hat{\boldsymbol{y}}) + R(\hat{\boldsymbol{z}}) + \lambda D(\boldsymbol{x}, \hat{\boldsymbol{x}}) \\
  &= \mathbb{E} \left[ -\log_2 \left( p_{\hat{\boldsymbol{y}}|\hat{\boldsymbol{z}}}(\hat{\boldsymbol{y}} \mid \hat{\boldsymbol{z}}) \right) \right] \\
  &\quad + \mathbb{E} \left[ -\log_2 \left( p_{\hat{\boldsymbol{z}}|\boldsymbol{\psi}}(\hat{\boldsymbol{z}} \mid \boldsymbol{\psi}) \right) \right] \\
  &\quad + \lambda D(\boldsymbol{x}, \hat{\boldsymbol{x}}),
\end{split}
\label{eq:eq8}
\end{equation}
where \(\lambda > 0\) controls the trade-off between bitrate and reconstruction quality. The terms \(R(\hat{\boldsymbol{y}})\) and \(R(\hat{\boldsymbol{z}})\) denote the estimated bitrates of \(\hat{\boldsymbol{y}}\) and \(\hat{\boldsymbol{z}}\), respectively, and \(D(\boldsymbol{x}, \hat{\boldsymbol{x}})\) measures the distortion between the input image \(\boldsymbol{x}\) and the reconstructed image \(\hat{\boldsymbol{x}}\).

In HCFSSNet, the main transforms \(g_a\) and \(g_s\), together with the hyperprior transforms \(h_a\), \(h_{\text{mean}}\), and \(h_{\text{scale}}\), are built from \(3\times3\) convolutions, sub-pixel \(3\times3\) convolutions, residual blocks with stride (RBS), residual block upsampling (RBU) layers, and the proposed HCFSS block. The RBS, RBU, and sub-pixel convolution layers follow the design of~\cite{cheng2020image}. The detailed designs of the HCFSS block, the Vision Frequency State Space (VFSS) block, and the frequency-aware entropy refinement module are described in the following subsections.

\subsection{Hybrid Convolution--Frequency State Space Block}
\label{sec:HCFSS_block}

Previous learned image compression methods have mainly relied on CNNs, Transformers, or their hybrids as nonlinear transforms. CNNs provide a strong local inductive bias, while sequence-based models are better suited for capturing broader contextual dependencies. More recently, state space models (SSMs) have emerged as an efficient alternative for long-range modeling in vision. In particular, modern selective SSMs such as Mamba~\cite{gu2024mamba} offer linear-time sequence modeling, making them attractive for high-resolution visual inputs. Motivated by the complementary properties of convolution and state-space modeling, we design a compact Hybrid Convolution--Frequency State Space (HCFSS) block for learned image compression.

As illustrated in Fig.~\ref{fig:subblocks}(a), the HCFSS block consists of a CNN branch for local representation refinement and a Vision Frequency State Space (VFSS) branch for complementary contextual modeling. This design combines the local modeling capability of convolutions with the efficient long-range dependency modeling of visual SSMs, while avoiding the quadratic complexity of Transformer-based attention.

\begin{center}
\begin{minipage}{\textwidth}
\centering
\includegraphics[width=0.9\linewidth]{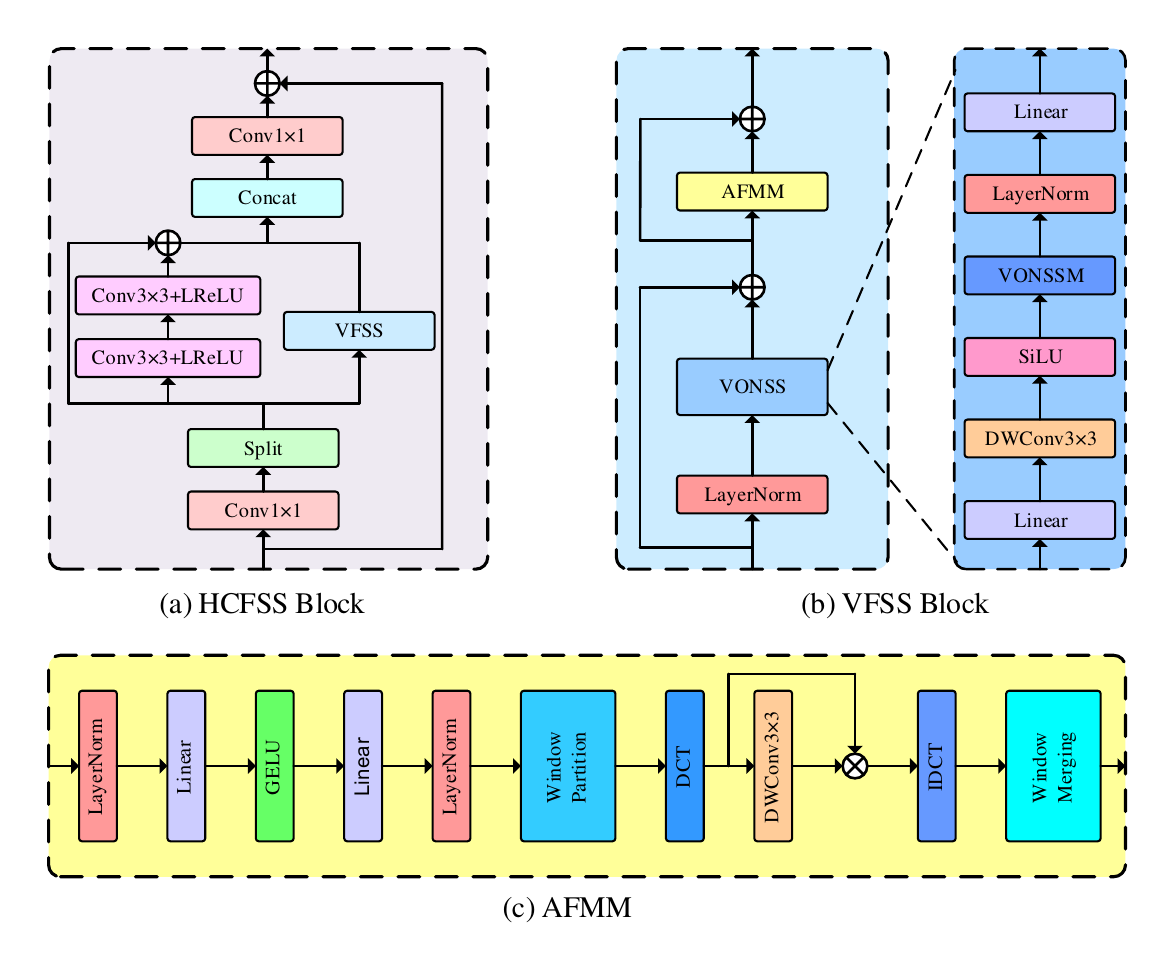}
\captionof{figure}{(a) Illustration of the proposed HCFSS block. LReLU denotes the Leaky ReLU activation function. ``Split'' and ``Concat'' refer to channel-wise feature separation and concatenation, respectively. VFSS denotes the Vision Frequency State Space block. (b) Architecture of the VFSS block. DwConv denotes depthwise convolution. VONSSM denotes the Vision Omni-directional Neighborhood State Space Module, and AFMM stands for the Adaptive Frequency Modulation Module. (c) Architecture of the AFMM. DCT and IDCT denote the Discrete Cosine Transform and its inverse, respectively.}
\label{fig:subblocks}
\end{minipage}
\end{center}

Given an input feature map \(X \in \mathbb{R}^{C \times H \times W}\), the HCFSS block first applies a \(1 \times 1\) convolution and then splits the transformed feature along the channel dimension into two equal parts, denoted as \(X_{\text{loc}} \in \mathbb{R}^{\frac{C}{2} \times H \times W}\) and \(X_{\text{ctx}} \in \mathbb{R}^{\frac{C}{2} \times H \times W}\):
\begin{equation}
\begin{aligned}
X_{\text{loc}}, X_{\text{ctx}} = \text{Split}(\text{Conv}_{1 \times 1}(X)).
\end{aligned}
\end{equation}

The local branch \(X_{\text{loc}}\) is processed by a CNN-based residual network \(\text{Res}(\cdot)\) to refine local spatial details:
\begin{equation}
\begin{aligned}
{X}'_{\text{loc}} = \text{Res}(X_{\text{loc}}).
\end{aligned}
\end{equation}

In parallel, the contextual branch \(X_{\text{ctx}}\) is fed into the VFSS block to capture complementary long-range contextual information:
\begin{equation}
\begin{aligned}
{X}'_{\text{ctx}} = \text{VFSS}(X_{\text{ctx}}).
\end{aligned}
\end{equation}

The outputs of the two branches are then concatenated and projected back to \(C\) channels by another \(1 \times 1\) convolution:
\begin{equation}
\begin{aligned}
X_{\text{fuse}} = \text{Conv}_{1 \times 1}\bigl(\text{Cat}({X}'_{\text{loc}}, {X}'_{\text{ctx}})\bigr).
\end{aligned}
\end{equation}

Finally, a residual connection is added to form the output feature map:
\begin{equation}
\begin{aligned}
X_{\text{output}} = X + X_{\text{fuse}}.
\end{aligned}
\end{equation}

Overall, the HCFSS block provides a compact way to combine local convolutional refinement with complementary contextual aggregation, making it suitable for both the main transform and the hyperprior-related pathways in HCFSSNet.

\subsection{Vision Frequency State Space Block}

To capture long-range dependencies in two-dimensional feature maps while introducing frequency-aware modulation, we design the Vision Frequency State Space (VFSS) block, which consists of a Vision Omni-directional Neighborhood State Space (VONSS) block and an Adaptive Frequency Modulation Module (AFMM).

A key challenge in applying state-space models (SSMs) to 2D feature maps lies in the mismatch between sequential scan distance and spatial neighborhood relations. Under conventional scan-based flattening, the sequential distance between two pixels does not necessarily reflect their Euclidean proximity, i.e., $\Delta_{\text{seq}} \neq \Delta_{\text{spatial}}$. As illustrated in Fig.~\ref{fig:motivation}, spatially adjacent pixels may be separated by long paths in the scan sequence, which weakens local dependency modeling.

\begin{center}
\begin{minipage}{\textwidth}
\centering
\includegraphics[width=\linewidth, trim=0cm 2.5cm 0cm 2cm,clip]{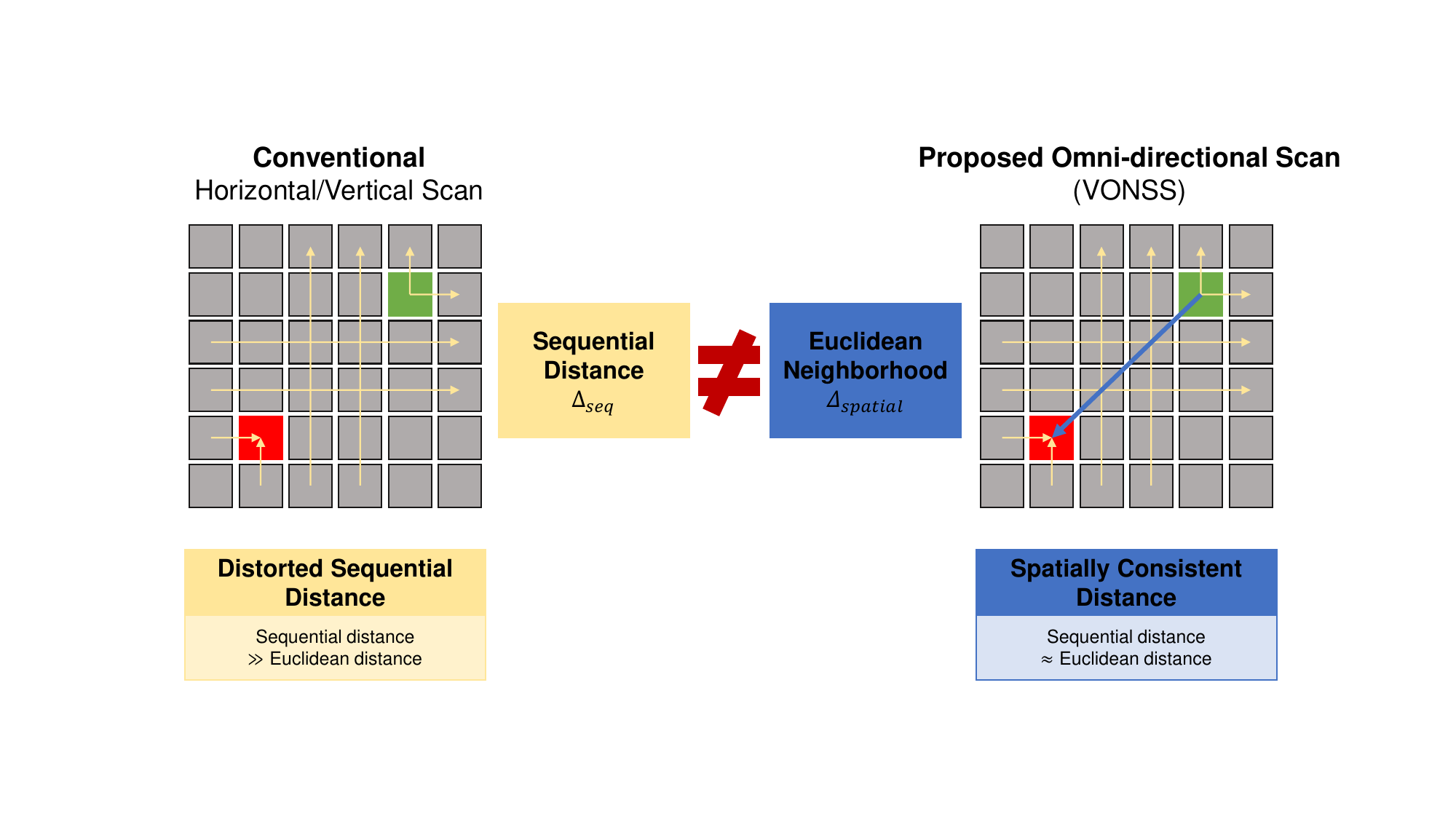}
\captionof{figure}{Comparison between conventional scan-based flattening and the proposed omni-directional scan. In conventional scans, the sequential distance between pixels does not correspond to their Euclidean proximity, leading to distorted local dependencies (left). In contrast, the proposed omni-directional scan aligns sequential distance with spatial neighborhood, enabling spatially consistent modeling in SSMs (right).}
\label{fig:motivation}
\end{minipage}
\end{center}

To address this issue, the VFSS block first applies VONSS to improve directional coverage of 2D neighborhood relations and then uses AFMM to introduce frequency-aware refinement. As shown in Fig.~\ref{fig:subblocks}(b), given an input feature map \(X_{\text{ctx}}\), the VFSS block first applies layer normalization and then processes the feature with VONSS to obtain a context-enhanced representation \(X_{\text{VONSS}}\). AFMM then performs adaptive frequency-domain reweighting on \(X_{\text{VONSS}}\) to produce the output of the VFSS block. The overall computation is written as
\begin{equation}
\begin{aligned}
X_{\text{VONSS}} &= \text{VONSS}(\text{LN}(X_{\text{ctx}})) + X_{\text{ctx}}, \\
X'_{\text{ctx}} &= \text{AFMM}(X_{\text{VONSS}}) + X_{\text{VONSS}}.
\end{aligned}
\end{equation}

\subsubsection{VONSS Block}

State-space models (SSMs) are originally designed for one-dimensional sequences. When extending them to image features, a common strategy is to convert a two-dimensional feature map into one or more 1D scan sequences. However, as discussed in Fig.~\ref{fig:motivation}, conventional scan-based flattening introduces a mismatch between sequential distance and spatial neighborhood relations. In particular, pixels that are adjacent in 2D space may become relatively distant along a given scan path, making some local dependencies less explicitly represented in the sequential domain. Existing visual SSM variants mainly rely on horizontal and vertical scan patterns. For example, Vim~\cite{Zhu2024vim} employs horizontal scans with two directions, while VMamba~\cite{liu2024vmamba} further extends directional scanning to both horizontal and vertical orders. Although these designs provide effective directional modeling, diagonal neighborhood relations are still less explicitly covered.

To better adapt SSMs to 2D image features, we propose the \textbf{Vision Omni-directional Neighborhood State Space (VONSS)} block. The key idea is to extend the scan design from conventional horizontal/vertical directions to a richer set of neighborhood-aware directions. Specifically, VONSS incorporates four scan families---horizontal, vertical, diagonal, and anti-diagonal---together with their reverse orders, resulting in eight scan directions in total. In this way, VONSS provides more complete directional coverage of first-order neighborhood relations while retaining the long-range modeling capability of selective state space models.

As illustrated in Fig.~\ref{fig:subblocks}(b), the VONSS block consists of linear layers, a depthwise convolution layer, a SiLU activation, the Vision Omni-directional Neighborhood State Space Module (VONSSM), and LayerNorm. Given an input feature map \(X_{\text{ln}}\), the processing flow is expressed as
\begin{equation}
\begin{aligned}
X^{\text{in}}_{\text{VONSSM}} &= \text{SiLU}\bigl(\text{DwConv}_{3 \times 3}(\text{Linear}(X_{\text{ln}}))\bigr), \\
X^{\text{out}}_{\text{VONSSM}} &= \text{VONSSM}(X^{\text{in}}_{\text{VONSSM}}), \\
X^{\text{out}}_{\text{VONSS}} &= \text{Linear}(\text{LN}(X^{\text{out}}_{\text{VONSSM}})).
\end{aligned}
\end{equation}

\begin{center}
\begin{minipage}{\textwidth}
\centering
\includegraphics[width=\linewidth]{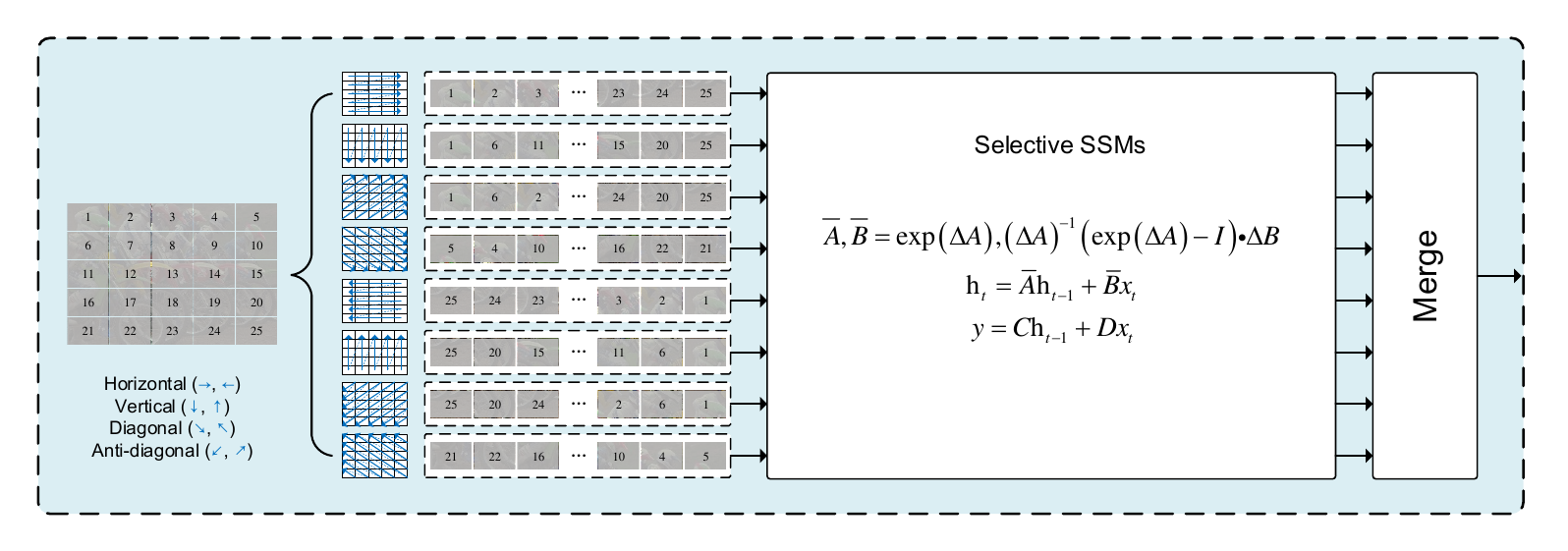}
\captionof{figure}{
Implementation overview of the proposed VONSSM. The input feature map is converted into multiple scan sequences along horizontal, vertical, diagonal, and anti-diagonal directions (including their reverse orders). These sequences are processed by selective state space models and then merged back into a 2D feature map.
}
\label{fig:4}
\end{minipage}
\end{center}

Figure~\ref{fig:4} shows the implementation overview of VONSSM. For a given input feature map \(X^{\text{in}}_{\text{VONSSM}}\), VONSSM first constructs eight directional scan sequences according to the proposed omni-directional scan scheme. These scan sequences are then processed by selective state space models~\cite{gu2024mamba} in parallel to capture directional contextual dependencies. Finally, the outputs from different scan directions are merged and reshaped back into a 2D feature map. By combining multiple scan orientations within a unified sequence-processing framework, VONSS enables spatial dependencies from different directions to be modeled more explicitly while remaining compatible with efficient SSM-based computation.

\subsubsection{Adaptive Frequency Modulation Module}

\begin{center}
\begin{minipage}{\textwidth}
\centering
\includegraphics[width=\linewidth, trim=0cm 0cm 0cm 4cm,clip]{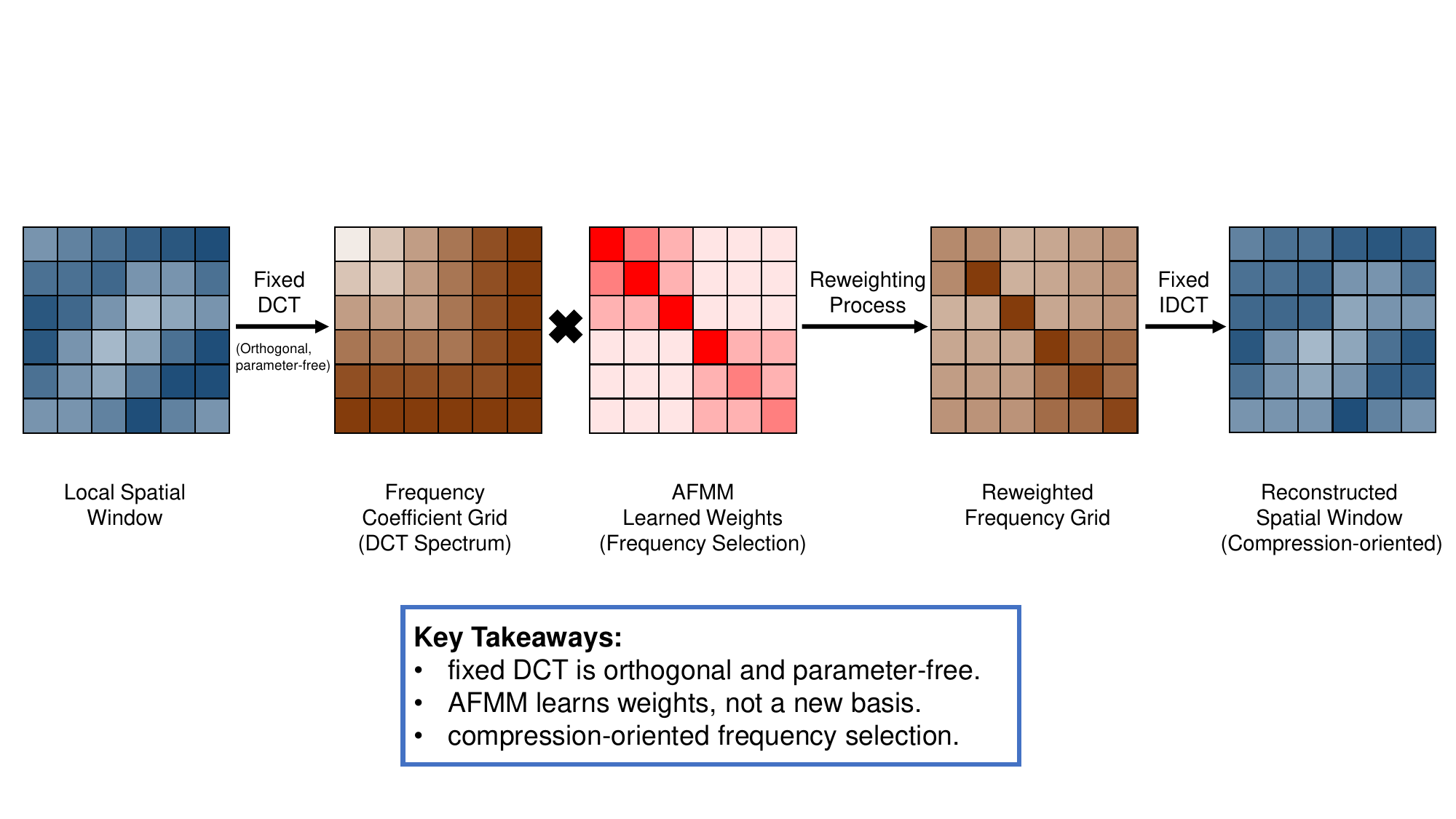}
\captionof{figure}{
Motivation of the proposed AFMM. A local spatial window is transformed into the frequency domain using a fixed DCT, which is orthogonal and parameter-free. Instead of learning a new frequency basis, AFMM performs element-wise reweighting on DCT coefficients, enabling compression-oriented frequency selection. The refined representation is then transformed back to the spatial domain via IDCT.
}
\label{fig:afmm_motivation}
\end{minipage}
\end{center}

We design an Adaptive Frequency Modulation Module (AFMM) to introduce frequency-aware refinement into the VFSS branch. Rather than explicitly splitting features into coarse high- and low-frequency groups, AFMM operates on windowed discrete cosine transform (DCT) coefficients and performs adaptive reweighting over a finer set of frequency components. As illustrated in Fig.~\ref{fig:afmm_motivation}, a local spatial window is first mapped to a fixed DCT domain, reweighted by AFMM, and then transformed back to the spatial domain. We adopt the DCT as a fixed transform because it is orthogonal, parameter-free, and lightweight, while providing a stable frequency-domain basis that is well aligned with transform coding in image compression. Importantly, AFMM learns adaptive weights on DCT coefficients rather than learning a new frequency basis, which keeps the module lightweight and stable while remaining easy to integrate into an existing compression pipeline.

As illustrated in Fig.~\ref{fig:subblocks}(c), given an input feature map \(X^{\text{in}}_{\text{AFMM}}\), AFMM first applies layer normalization and two linear layers with a GELU activation in between to adjust the channel representation:
\begin{equation}
\begin{aligned}
X^{\text{act}}_{\text{AFMM}} &= \text{LN}\bigl(\text{Linear}(\text{GELU}(\text{Linear}(\text{LN}(X^{\text{in}}_{\text{AFMM}}))))\bigr).
\end{aligned}
\end{equation}

Next, the feature map is partitioned into non-overlapping windows of size \(w \times w\) using the window-partition operator \(\text{Wp}(\cdot)\). For each window, the DCT is applied to obtain frequency-domain coefficients, and a \(3 \times 3\) depthwise convolution is used to estimate adaptive modulation weights for different frequency responses:
\begin{equation}
\begin{aligned}
F &= \text{DCT}(\text{Wp}(X^{\text{act}}_{\text{AFMM}})), \\
W &= \text{Conv}_{3 \times 3}(F).
\end{aligned}
\end{equation}

The estimated weights \(W\) are then applied to the frequency coefficients by element-wise multiplication. The modulated coefficients are transformed back to the spatial domain by inverse DCT (IDCT) and merged by the window-merging operator \(\text{Wm}(\cdot)\):
\begin{equation}
\begin{aligned}
X^{\text{freq}}_{\text{AFMM}} &= \text{Wm}(\text{IDCT}(W \odot F)),
\end{aligned}
\end{equation}
where \(\odot\) denotes element-wise multiplication. In this way, AFMM provides frequency-aware refinement of the feature representation through adaptive reweighting in the DCT domain. The contributions of AFMM and VONSS are further examined in the ablation studies in Section~\ref{sec:ablation}.

\subsection{Channel-wise Entropy Model}

We further introduce a frequency-aware refinement design into the channel-wise entropy model. The overall entropy model follows the standard hyperprior-plus-slice framework, while the main modification lies in how side information is processed before predicting the Gaussian parameters of each latent slice. The architecture is illustrated in Fig.~\ref{fig:6}.

Given the latent representation \(\boldsymbol{y}\), the hyper-analysis transform \(h_a\) first maps it to the hyperprior \(\boldsymbol{z}\), which is entropy-coded to generate the side information. After decoding, the reconstructed hyperprior \(\hat{\boldsymbol{z}}\) is fed into two hyper-synthesis transforms, \(h_{\text{mean}}\) and \(h_{\text{scale}}\), to produce the mean-related features \(F_{\text{mean}}\) and scale-related features \(F_{\text{scale}}\), respectively. In our implementation, \(h_a\), \(h_{\text{mean}}\), and \(h_{\text{scale}}\) are built from convolutional layers and HCFSS blocks, so that the hyperprior pathway can incorporate both local refinement and contextual modeling.

For the \(i\)-th channel slice, the slice network \(e_i\) takes as input the mean features \(F_{\text{mean}}\), scale features \(F_{\text{scale}}\), the current latent slice \(y_i\), and the previously reconstructed slices \(\{\bar{y}_0, \bar{y}_1, \ldots, \bar{y}_{i-1}\}\), and predicts the Gaussian parameters \(\mu_i\) and \(\sigma_i\) for entropy coding. Since \(F_{\text{mean}}\) and \(F_{\text{scale}}\) have relatively low spatial resolution, modeling them does not require the same long-sequence processing capacity as the main transform. In this setting, window-based self-attention provides an effective way to refine nearby contextual interactions, offering more direct local context aggregation than another long-sequence SSM while remaining more expressive than a purely convolutional refinement module. We therefore combine a Swin Transformer block with AFMM to form a Frequency Swin Transformer Attention Module (FSTAM), which introduces frequency-aware modulation into hyperprior feature refinement within each window.

With the predicted parameters \(\mu_i\) and \(\sigma_i\), each channel slice \(y_i\) is encoded using the corresponding Gaussian model. In this way, the proposed entropy model preserves the standard channel-slice coding framework while incorporating frequency-aware refinement in the side-information pathway.

\begin{center}
\begin{minipage}{\textwidth}
\centering
\includegraphics[width=0.9\linewidth]{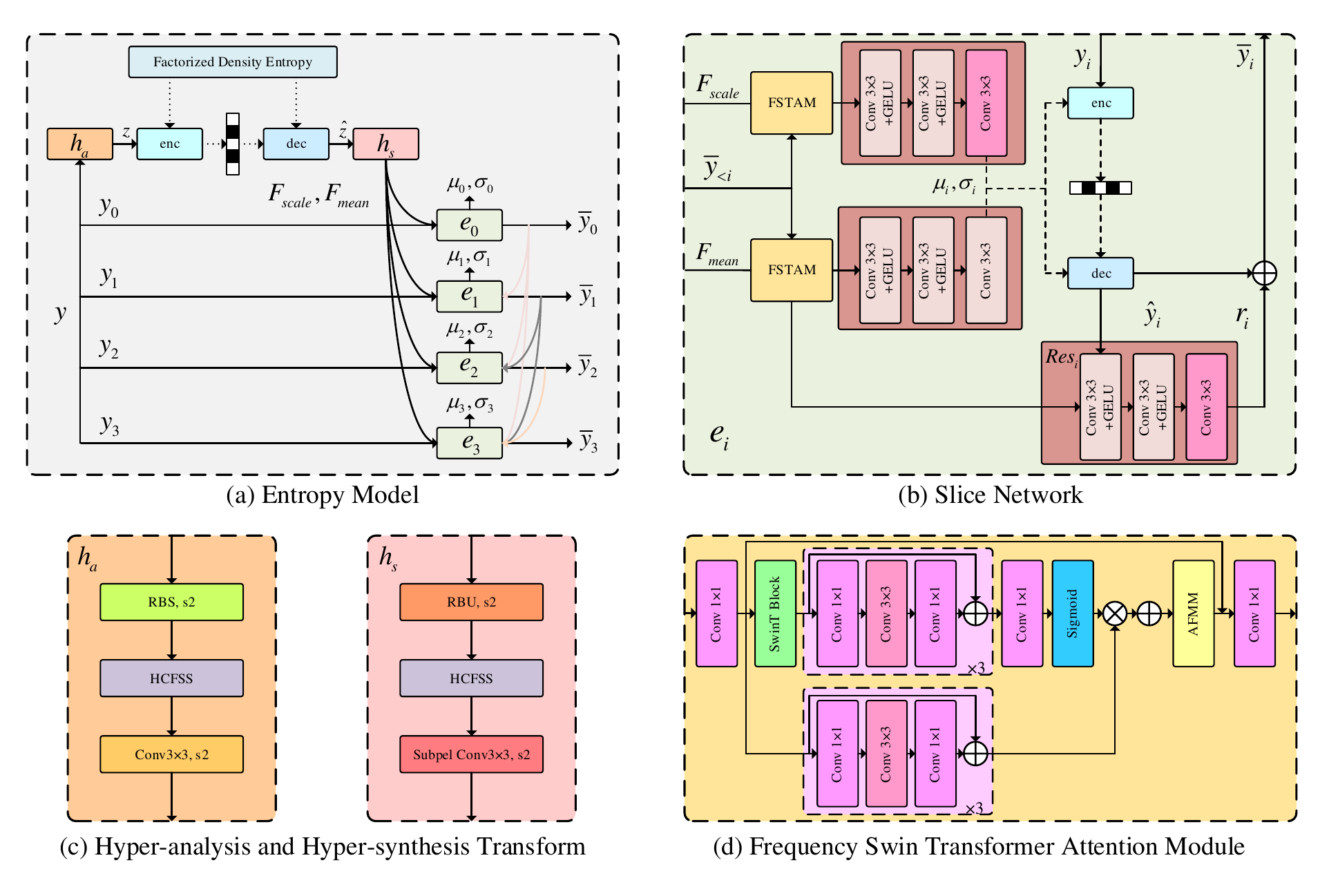}
\captionof{figure}{Architectural details of the proposed channel-wise entropy model. 
(a) Overall structure of the entropy model. 
(b) Slice network \(e_i\). 
(c) Hyper-analysis and hyper-synthesis transforms. 
(d) Frequency Swin Transformer Attention Module (FSTAM); ``SwinT block'' denotes a Swin Transformer block.}
\label{fig:6}
\end{minipage}
\end{center}

\section{Experiments}

\subsection{Experimental Setup}

\subsubsection{Training Configuration}
We train HCFSSNet in two stages. The model is first pre-trained on 300K images from ImageNet~\cite{deng2009imagenet} and then fine-tuned on 80K images from LSDIR~\cite{li2023lsdir}. During fine-tuning on LSDIR, each training image is randomly cropped to \(512 \times 512\) pixels, and the global batch size is set to 2. The initial learning rate is set to \(1 \times 10^{-4}\), and decayed by a factor of 10 at approximately 0.4M and 0.6M iterations.

Training follows the rate--distortion objective in Eq.~\ref{eq:eq8}, where mean-squared error (MSE) is used as the distortion term. The Lagrange multiplier \(\lambda\) is set to \(\{0.0025, 0.0035, 0.0067, 0.0130, 0.0250, 0.0500\}\) to obtain models at different rate points. The window size of AFMM in the main HCFSS blocks is set to 16, while the FSTAM modules use a window size of 8. Unless otherwise specified, the remaining training settings follow~\cite{liu2023learned}. All models are trained on two NVIDIA RTX 4090 GPUs and evaluated on a single NVIDIA RTX 4090 GPU.

\subsubsection{Evaluation Protocol}
We evaluate the proposed HCFSSNet on three public image compression benchmarks: Kodak (24 images, \(768 \times 512\) pixels)~\cite{kodak1993kodak}, Tecnick (100 images, \(1200 \times 1200\) pixels)~\cite{asuni2014testimages}, and the CLIC Professional Validation set (41 images, approximately 2K resolution)~\cite{toderici2020clic}. Reconstruction quality is measured by peak signal-to-noise ratio (PSNR), and bitrate is reported in bits per pixel (bpp). For overall rate--distortion comparison, we report BD-rate results with VTM as the anchor.

\subsubsection{Model Configuration}
Unless otherwise specified, the channel width in each HCFSS block is set to 256, with 128 channels allocated to the convolutional branch and 128 channels allocated to the VFSS branch. The latent representation \(\boldsymbol{y}\) uses 320 channels, and the hyper-latent \(\boldsymbol{z}\) uses 192 channels. The hyper-analysis and hyper-synthesis transforms are configured with 256 channels. For entropy coding, we adopt a five-slice channel-wise entropy model.

\subsection{Rate--Distortion Performance}

We compare HCFSSNet with the conventional VTM anchor and several recent learned image compression codecs, including InvCompress~\cite{xie2021enhanced}, MLIC++~\cite{jiang2023mlicpp}, TCM~\cite{liu2023learned}, WeConvene~\cite{Fu2024ECCV2024}, CCA~\cite{han2024causal}, FTIC~\cite{li2024frequencyaware}, and MambaIC~\cite{zeng2025mambaic}. Figures~\ref{fig:rd_results}(a)--(c) show the PSNR--bitrate curves on Kodak, Tecnick, and the CLIC Professional Validation set, respectively. BD-rate is computed from the PSNR--bitrate curves using the Bjøntegaard method over the common quality range.

Compared with the VTM anchor, HCFSSNet consistently achieves BD-rate savings on all three datasets, with reductions of 18.06\% on Kodak, 24.56\% on Tecnick, and 22.44\% on CLIC. As shown in Fig.~\ref{fig:rd_results}, the proposed model remains competitive across the evaluated bitrate range on all three benchmarks.


\begin{center}
\begin{minipage}{\textwidth}
\centering
\begin{minipage}[t]{0.3\textwidth}
\centering
\includegraphics[width=\linewidth]{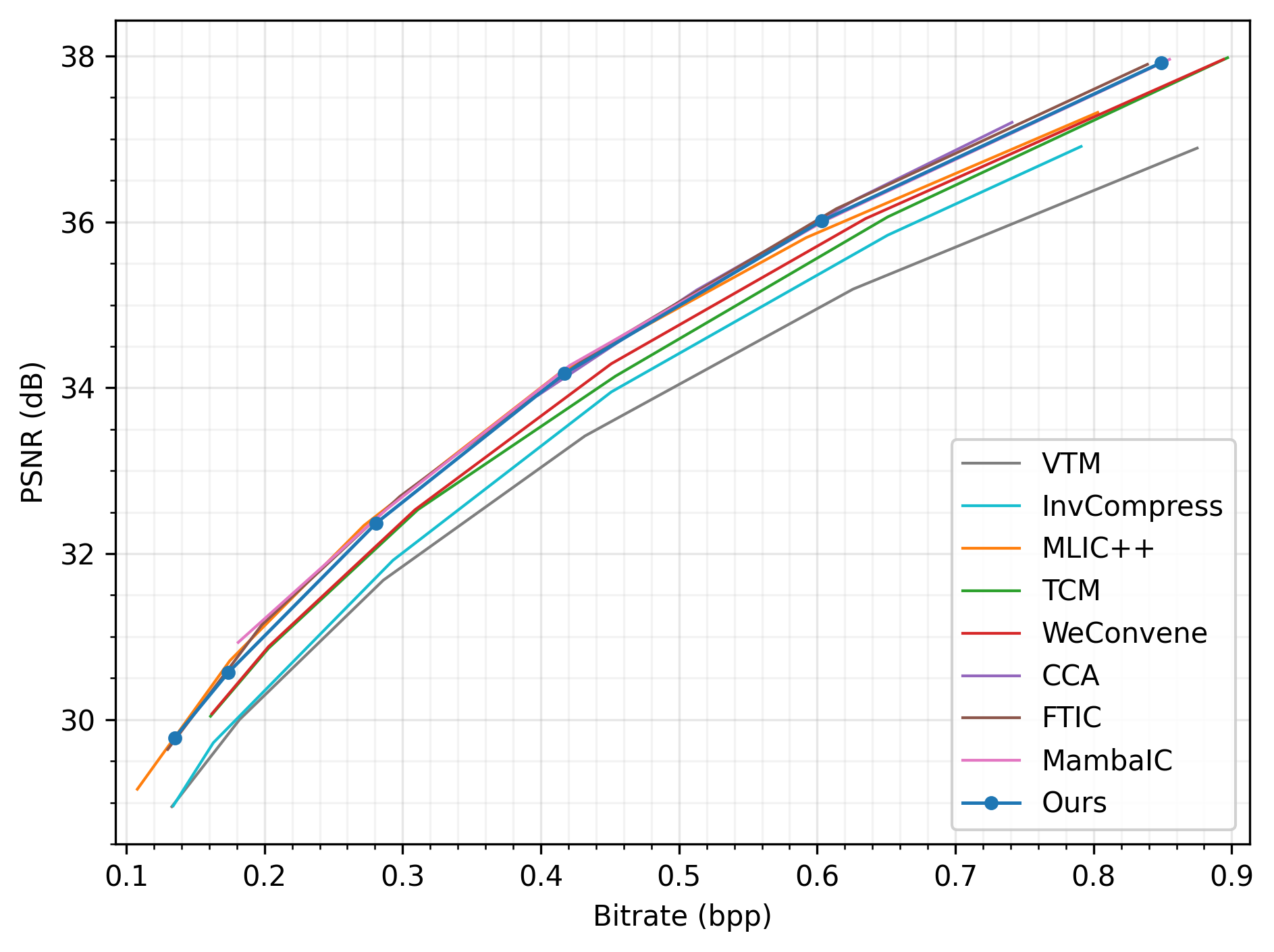}\\
(a)
\end{minipage}\hfill
\begin{minipage}[t]{0.3\textwidth}
\centering
\includegraphics[width=\linewidth]{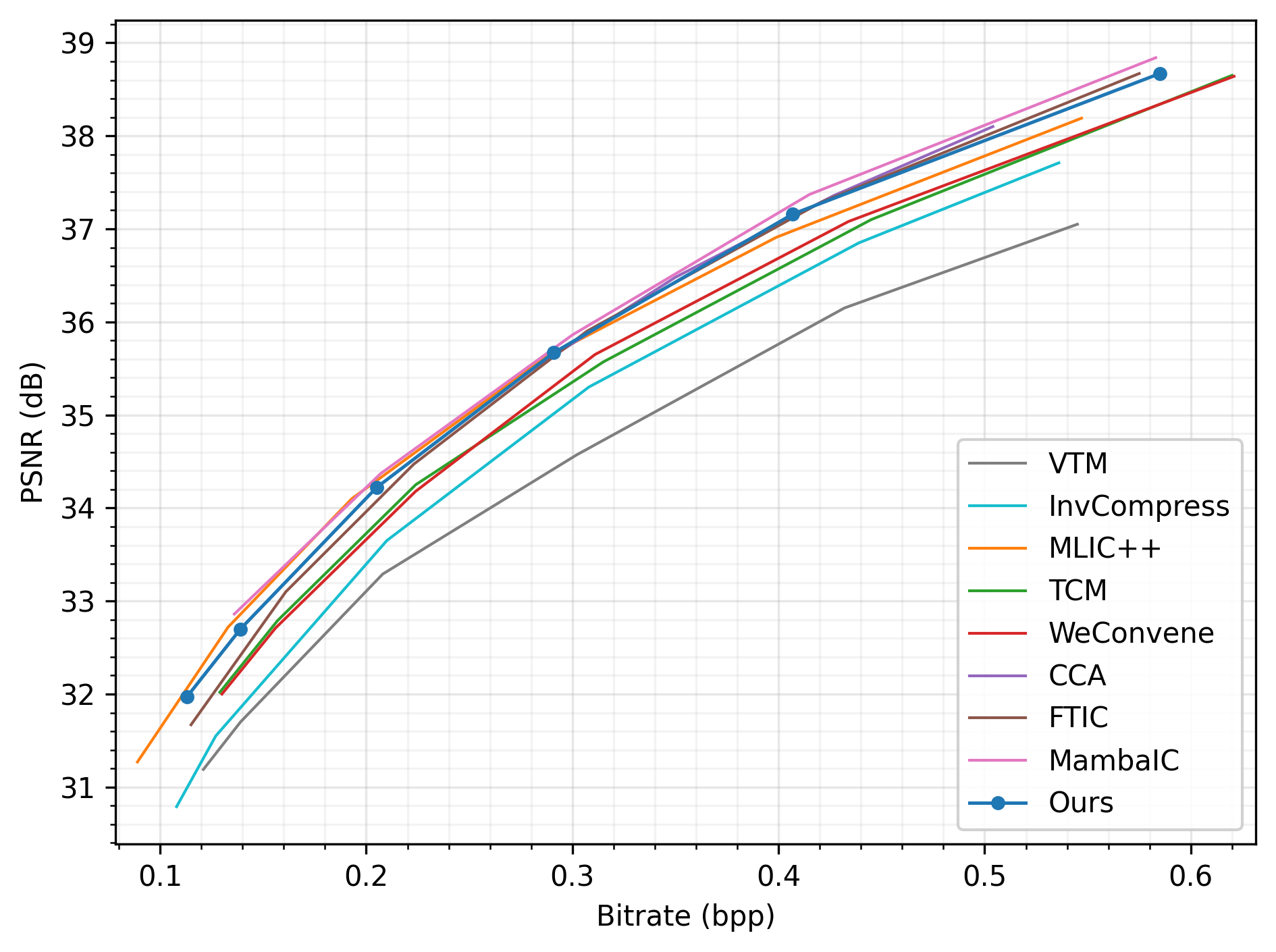}\\
(b)
\end{minipage}\hfill
\begin{minipage}[t]{0.3\textwidth}
\centering
\includegraphics[width=\linewidth]{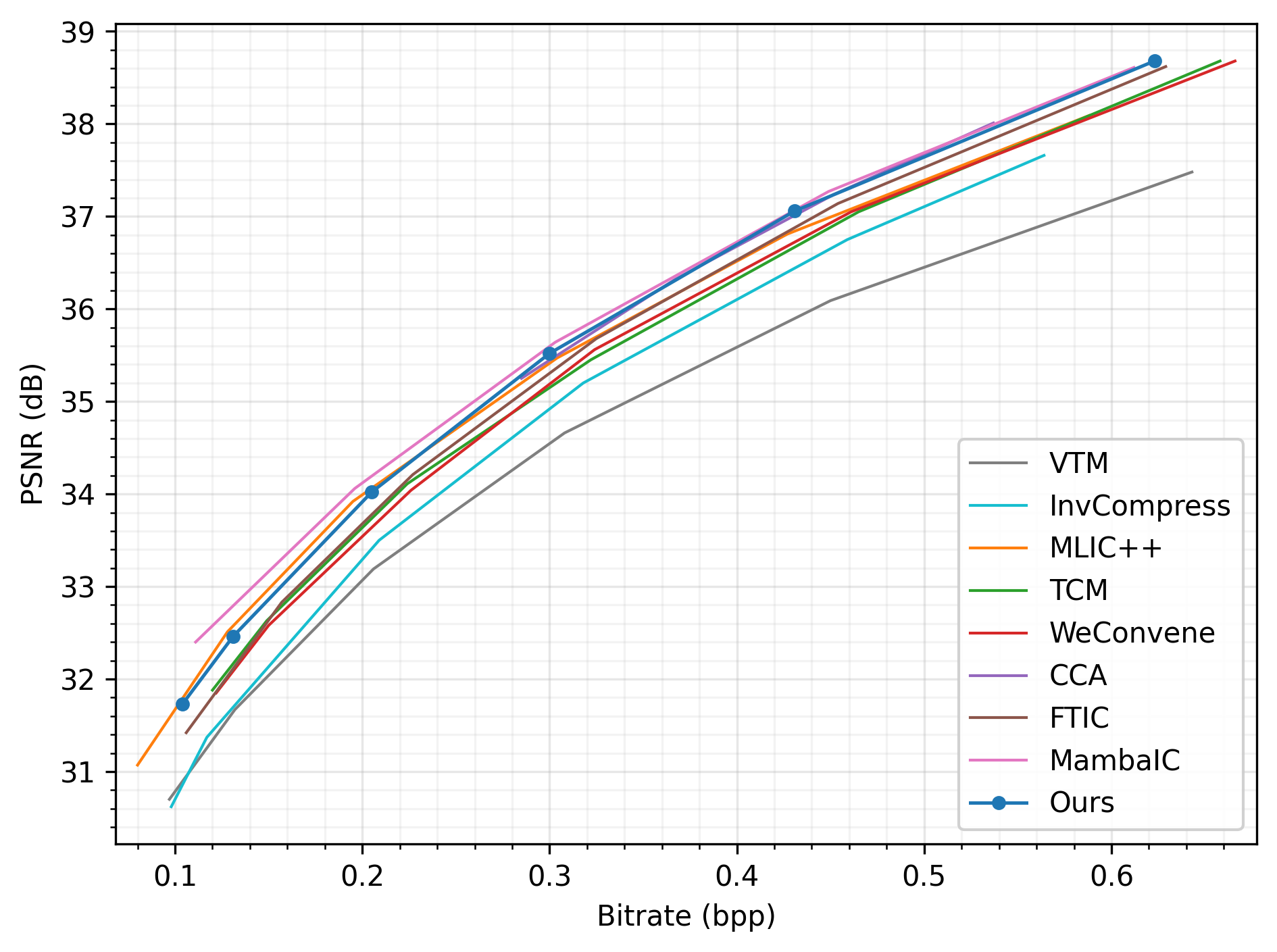}\\
(c)
\end{minipage}

\captionof{figure}{Rate--distortion results (PSNR vs. bitrate, in bpp). From left to right: (a) Kodak, (b) Tecnick, and (c) CLIC Professional Validation. The curves compare state-of-the-art learned image compression methods with the conventional VTM codec; higher PSNR at the same bitrate (or lower bitrate at the same PSNR) indicates better performance.}
\label{fig:rd_results}
\end{minipage}
\end{center}

Table~\ref{tab:rd_results_wide} further summarizes the model complexity, decoding time, and BD-rate comparisons with recent learned codecs. HCFSSNet achieves competitive rate--distortion performance relative to recent CNN-, Transformer-, and SSM-based models, while keeping the model size below several high-capacity baselines. In particular, HCFSSNet uses 80.97M parameters, compared with 116.72M for MLIC++ and 123.81M for MambaIC. In terms of BD-rate, HCFSSNet is close to these stronger baselines but does not outperform them on all datasets: MLIC++ achieves lower BD-rate on Kodak and Tecnick, and MambaIC achieves lower BD-rate on all three benchmarks. We therefore position HCFSSNet as a compact and competitive design rather than a new state of the art, with its main trade-off being a moderate model size and competitive compression performance rather than minimum runtime.

\begin{center}
\begin{minipage}{\textwidth}
\centering
\includegraphics[width=0.7\linewidth]{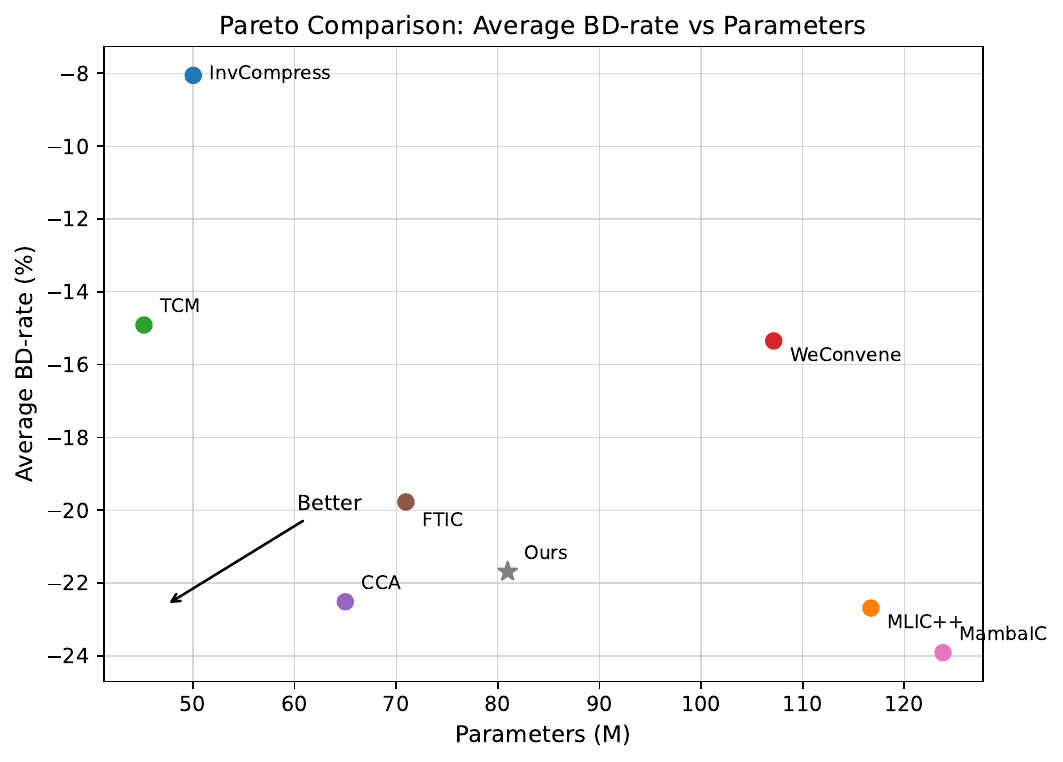}
\captionof{figure}{Average BD-rate versus parameter count for recent learned image compression methods. Lower is better in both dimensions. HCFSSNet lies in a compact-but-competitive region, using fewer parameters than several high-capacity baselines while maintaining competitive average BD-rate.}
\label{fig:pareto_params}
\end{minipage}
\end{center}

To better illustrate the trade-off between compression performance and model size, we further provide a comparison of average BD-rate versus parameter count in Fig.~\ref{fig:pareto_params}. HCFSSNet lies in a compact-but-competitive region of the current LIC design space: it uses fewer parameters than several high-capacity baselines such as MLIC++ and MambaIC, while maintaining competitive average BD-rate.

Compared with FTIC~\cite{li2024frequencyaware}, which also introduces frequency-aware modeling, HCFSSNet obtains lower BD-rate on the higher-resolution Tecnick and CLIC datasets, with gains of 2.29\% and 4.54\%, respectively, while the Kodak result is slightly weaker. This comparison suggests that the proposed hybrid design remains competitive on higher-resolution benchmarks, although its advantage is not uniform across all datasets.

In terms of decoding efficiency, HCFSSNet is faster than the VTM anchor, but it is not Pareto-optimal in runtime among the compared learned codecs in Table~\ref{tab:rd_results_wide}. Its decoding time is higher than those of MLIC++, TCM, WeConvene, CCA, FTIC, and MambaIC. This additional runtime is mainly associated with the omni-directional scanning and frequency-aware refinement modules introduced in the main and hyperprior pathways. Therefore, the main trade-off of HCFSSNet is not minimum latency, but rather a compact model size together with competitive rate--distortion performance.

Overall, the results indicate that HCFSSNet occupies a middle point in the current LIC design space: it does not reach the best BD-rate or the best runtime, but it provides a unified hybrid architecture that combines convolution-based local refinement, state-space-based contextual modeling, and frequency-aware modulation within a moderate parameter budget.

\begin{table*}[t]
  \centering
  \caption{Model complexity, decoding time, and BD-rate (\%) comparison between the conventional VTM codec and previous learned image compression methods. BD-rate is measured relative to VTM-23.13 (negative values indicate bitrate savings over VTM).}
  \label{tab:rd_results_wide}
  \footnotesize
  \setlength{\tabcolsep}{4pt}
  \begin{tabular}{
      l
      S[table-format=3.2]
      S[table-format=2.3]
      S[table-format=-2.2]
      S[table-format=-2.2]
      S[table-format=-2.2]
  }
    \toprule
    \multicolumn{1}{c}{Method} &
    \multicolumn{1}{c}{Params (M)} &
    \multicolumn{1}{c}{Dec. (s/img)} &
    \multicolumn{1}{c}{Kodak} &
    \multicolumn{1}{c}{Tecnick} &
    \multicolumn{1}{c}{CLIC} \\
    \midrule
    VTM-23.13 & {--} & 0.189 & 0.00 & 0.00 & 0.00 \\
    \addlinespace[2pt]\midrule\addlinespace[2pt]
    InvCompress~\cite{xie2021enhanced} & 50.03 & 1.758 & -5.43\% & -10.46\% & -8.28\% \\
    MLIC++~\cite{jiang2023mlicpp} & 116.72 & 0.021 & -19.24\% & -26.11\% & -22.71\% \\
    TCM~\cite{liu2023learned} & 45.18 & 0.035 & -11.60\% & -16.98\% & -16.16\% \\
    WeConvene~\cite{Fu2024ECCV2024} & 107.15 & 0.022 & -13.20\% & -17.17\% & -15.68\% \\
    CCA~\cite{han2024causal} & 64.99 & 0.063 & -18.30\% & -25.99\% & -23.25\% \\
    FTIC~\cite{li2024frequencyaware} & 70.96 & 0.052 & -19.15\% & -22.27\% & -17.90\% \\
    MambaIC~\cite{zeng2025mambaic} & 123.81 & 0.177 & -18.92\% & -26.84\% & -25.98\% \\
    Ours & 80.97 & 0.326 & -18.06\% & -24.56\% & -22.44\% \\
    \bottomrule
  \end{tabular}
\end{table*}

\subsection{Visual Results}

We further provide qualitative comparisons on the Tecnick dataset. Three representative images are selected and compressed by VTM, InvCompress, TCM, WeConvene, FTIC, and the proposed HCFSSNet. The reconstructed results are shown in Fig.~\ref{fig:results_vis_1}.

As illustrated in Fig.~\ref{fig:results_vis_1}, HCFSSNet tends to preserve clearer local structures and texture patterns in several representative regions, especially around thin edges, repeated textures, and fine line structures. Compared with the other methods shown here, the reconstructions produced by HCFSSNet exhibit fewer visible distortions in some challenging areas, while maintaining competitive bitrate and PSNR performance. These qualitative observations are broadly consistent with the quantitative comparisons reported in the previous subsection.


\noindent
\begin{minipage}{\textwidth}
\centering
\includegraphics[width=0.96\textwidth]{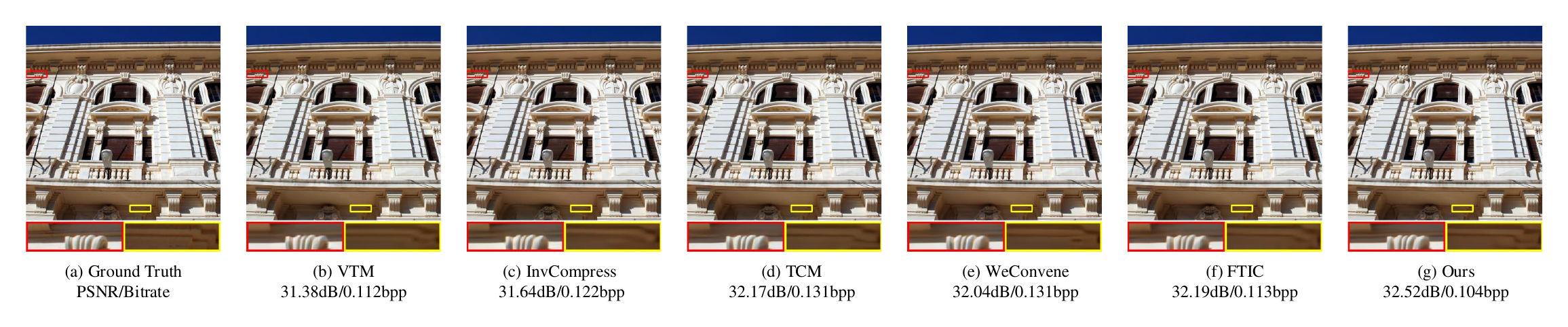}\vspace{-3mm}

\includegraphics[width=0.96\textwidth]{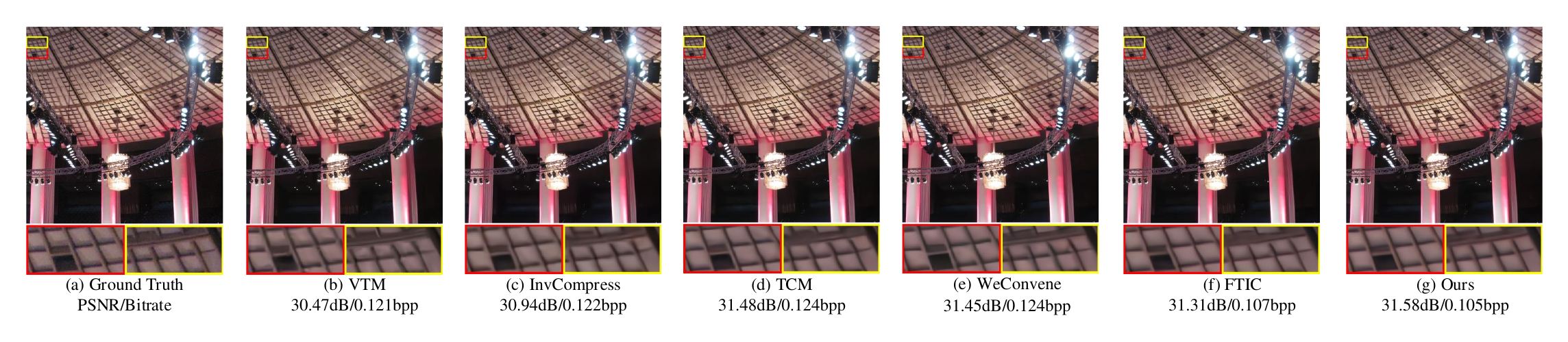}\vspace{-3mm}

\includegraphics[width=0.96\textwidth]{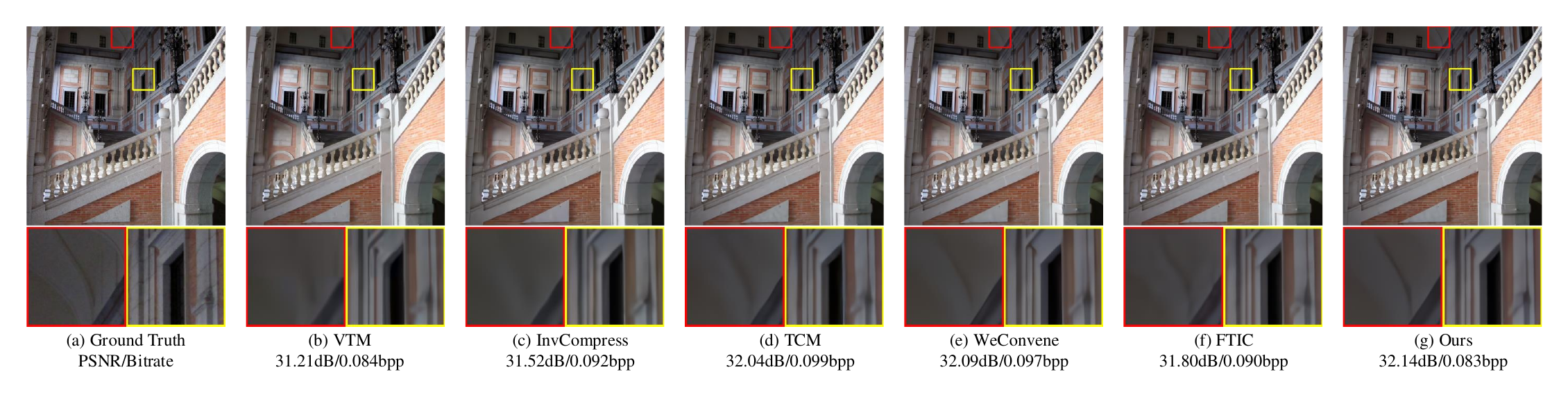}

\captionof{figure}{Visual comparisons on the Tecnick dataset~\cite{asuni2014testimages}. In the selected examples, HCFSSNet tends to preserve clearer local structures and texture patterns in several representative regions. Please zoom in for a clearer view.}
\label{fig:results_vis_1}
\end{minipage}

\subsection{Ablation Studies}
\label{sec:ablation}

To examine the contribution of the main components in HCFSSNet, we conduct ablation studies on VONSS, AFMM, and FSTAM. Unless otherwise specified, the ablation variants are trained on a randomly selected subset of 100K training images (with resolution \(\geq 256 \times 256\)) using a batch size of 8, and are evaluated on the Kodak dataset. For clarity, any experiment using a different training setup is stated explicitly in the corresponding subsection.

These ablations are designed to examine the necessity of three key design choices in HCFSSNet. Specifically, the comparison between CSM and VONSS addresses whether omni-directional scanning is beneficial for applying state-space modeling to 2D feature maps. The comparison with and without AFMM evaluates whether frequency-domain modulation brings measurable rate--distortion gains in the VFSS branch. Finally, the comparison between SWAtten and FSTAM examines whether frequency-aware refinement in the hyperprior pathway provides additional benefit beyond a standard Swin-based attention module.

\subsubsection{Scan-pattern ablation: VMamba-style Cross-Scan Module versus the proposed VONSS}

We first study the effect of the scan design in the proposed VONSS block. Specifically, we compare a variant using VONSS with another variant in which VONSS is replaced by the Cross-Scan Module (CSM) from VMamba~\cite{liu2024vmamba}. To focus on the scan design itself, both variants are trained without AFMM and FSTAM. For this comparison, both models are trained from scratch on 300K randomly selected images from ImageNet, each with a minimum resolution of \(256 \times 256\), and are evaluated on the Kodak dataset.

As shown in Fig.~\ref{fig:7}(a), the model equipped with VONSS achieves consistently higher PSNR than the CSM-based variant over the tested bitrate range. On the Kodak dataset, at three operating points around 0.45, 0.65, and 0.92 bpp, replacing CSM with VONSS improves PSNR from 34.11 to 34.16 dB, from 35.73 to 35.83 dB, and from 37.75 to 37.90 dB, respectively, while the bitrate changes only slightly (within 0.001--0.007 bpp).

These results suggest that extending the scan design from horizontal/vertical directions to omni-directional neighborhood coverage is beneficial for applying state-space modeling to 2D feature maps in learned image compression.

\subsubsection{Frequency-modulation ablation: w/o AFMM versus w/ AFMM}

We next study the effect of frequency-domain modulation by comparing the VONSS-based variant without AFMM and the full VFSS variant with AFMM. Starting from the VONSS-based variant, we add AFMM to form the full VFSS block and compare it with the counterpart without AFMM. As shown in Fig.~\ref{fig:7}(b), introducing AFMM further improves the rate--distortion performance over the tested bitrate range.

More specifically, on the Kodak dataset, at three operating points around 0.47, 0.69, and 1.01 bpp, adding AFMM improves PSNR from 33.64 to 33.73 dB, from 35.30 to 35.52 dB, and from 37.16 to 37.35 dB, respectively, while slightly reducing the bitrate from 0.475 to 0.465 bpp, from 0.688 to 0.682 bpp, and from 1.007 to 0.968 bpp. These results correspond to roughly 1--4\% bitrate reduction at similar reconstruction quality.

These observations suggest that introducing adaptive frequency-domain reweighting in the VFSS branch is beneficial for compression-oriented representation learning. In particular, AFMM provides a consistent improvement over the VONSS-based baseline on Kodak.

\subsubsection{Hyperprior-refinement ablation: SWAtten versus the proposed FSTAM}

Finally, we study the effect of hyperprior refinement by comparing FSTAM with the Swin-based attention module (SWAtten) used in the entropy model. We compare a variant using FSTAM with another variant in which FSTAM is replaced by the Swin-based attention module (SWAtten) from~\cite{liu2023learned}. As shown in Fig.~\ref{fig:7}(c), the FSTAM-based model achieves slightly better rate--distortion performance than the SWAtten-based counterpart over the tested bitrate range.

On the Kodak dataset, at two operating points around 0.68 and 0.97 bpp, replacing SWAtten with FSTAM improves PSNR from 35.52 to 35.56 dB and from 37.35 to 37.43 dB, respectively, while keeping the bitrate nearly unchanged (0.682 vs.\ 0.680 bpp and 0.968 vs.\ 0.969 bpp). Although the numerical gains are modest (about 0.04--0.08 dB), they are consistent across the tested operating points.

These results suggest that introducing frequency-aware refinement into the hyperprior pathway is beneficial for side-information modeling. In particular, combining a Swin Transformer block with AFMM provides a small but stable improvement over the SWAtten-based baseline on Kodak. Table~\ref{tab:ablation_summary} summarizes the main observations from Fig.~\ref{fig:7}: VONSS improves the scan design over CSM, AFMM brings consistent rate--distortion gains, and FSTAM provides small but stable improvements in hyperprior refinement.

\begin{center}
\begin{minipage}{\textwidth}
\centering
\includegraphics[width=\linewidth]{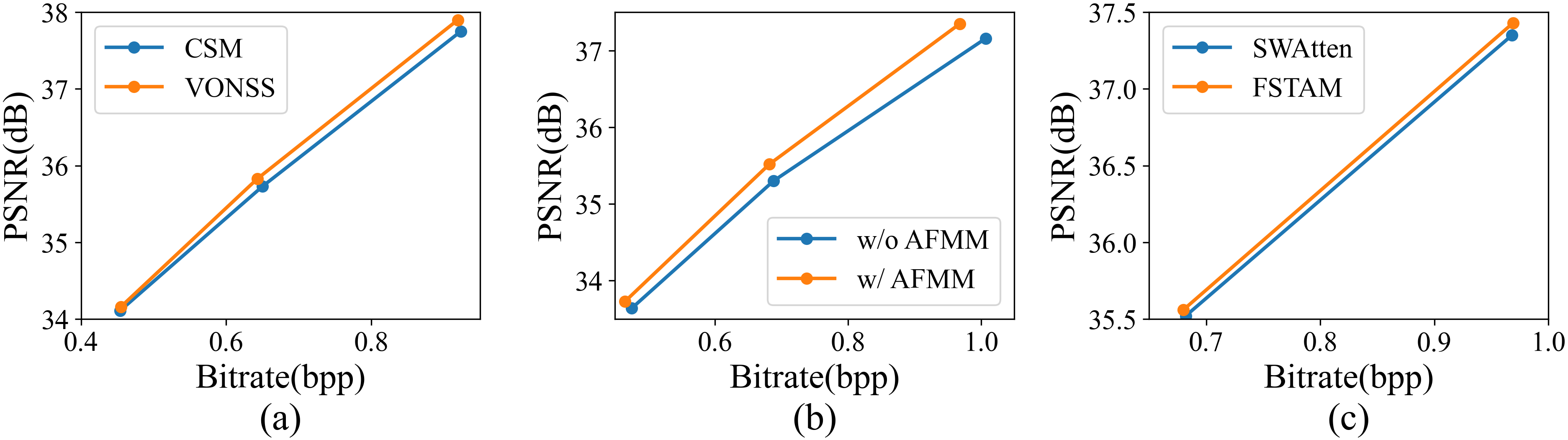}
\captionof{figure}{Ablation studies on the Kodak dataset. (a) Comparison between the Cross-Scan Module (CSM) and the proposed VONSS block. (b) Ablation of AFMM, where ``w/'' denotes ``with'' and ``w/o'' denotes ``without''. (c) Comparison between the Swin-based attention module (SWAtten) and the proposed FSTAM.}
\label{fig:7}
\end{minipage}
\end{center}

\begin{table*}[t]
\centering
\caption{Summary of ablation results on the Kodak dataset. Representative operating points are selected from Fig.~\ref{fig:7}. PSNR gain and bitrate change are computed as \emph{(proposed variant $-$ baseline)}. Negative bitrate change indicates bitrate reduction.}
\label{tab:ablation_summary}
\footnotesize
\setlength{\tabcolsep}{5pt}
\renewcommand{\arraystretch}{1.12}

\begin{tabular}{l l c c c}
\toprule
\textbf{Module} & \textbf{Baseline} & \textbf{Kodak bpp} & \textbf{PSNR gain (dB)} & \textbf{Bitrate change (bpp)} \\
\midrule
VONSS & CSM & 0.45 / 0.65 / 0.92 & +0.05 / +0.10 / +0.15 & +0.001 / $-0.007$ / $-0.004$ \\
AFMM & w/o AFMM & 0.47 / 0.69 / 1.01 & +0.09 / +0.22 / +0.19 & $-0.010$ / $-0.006$ / $-0.039$ \\
FSTAM & SWAtten & 0.68 / 0.97 & +0.04 / +0.08 & $-0.002$ / +0.001 \\
\bottomrule
\end{tabular}
\end{table*}

\subsection{Discussion}

It is worth noting that several recent LIC models, including MambaIC~\cite{zeng2025mambaic}, DCAE~\cite{lu2025learned}, and LALIC~\cite{feng2025linear}, report lower BD-rate than HCFSSNet on some benchmarks. These methods explore different design choices for improving compression performance, such as deeper state-space backbones, more sophisticated entropy modeling modules, or alternative linear-attention mechanisms.

In comparison, HCFSSNet is not designed to minimize BD-rate or runtime at all costs. As shown in Table~\ref{tab:rd_results_wide}, its main characteristic is a moderate model size together with a unified hybrid architecture that combines convolution-based local refinement, state-space-based contextual modeling, and frequency-aware modulation in both the main transform and the hyperprior pathway. With 80.97M parameters, HCFSSNet is smaller than several recent high-capacity baselines such as MambaIC and MLIC++, while maintaining competitive rate--distortion performance.

At the same time, HCFSSNet is not among the fastest learned codecs in our comparison. The additional runtime is mainly associated with the omni-directional scan design and the frequency-aware refinement modules introduced in the main and hyperprior transforms. Therefore, HCFSSNet should be viewed as a compact and competitive design point rather than a solution optimized for minimum latency.

\section{Conclusion}

In this work, we presented HCFSSNet, a compact hybrid convolution--frequency state space architecture for learned image compression. The proposed model integrates convolution-based local refinement, visual state space modeling, and frequency-aware modulation within a unified framework.

Experimental results on Kodak, Tecnick, and the CLIC Professional Validation set show that HCFSSNet consistently improves BD-rate over the conventional VTM anchor. In particular, HCFSSNet achieves BD-rate savings of 18.06\%, 24.56\%, and 22.44\% on Kodak, Tecnick, and CLIC, respectively, while using fewer parameters than several recent high-capacity baselines such as MambaIC and MLIC++. These results indicate that HCFSSNet provides a compact and competitive hybrid design for integrating local convolutional refinement, state space based contextual modeling, and frequency-aware modulation in learned image compression.

At the same time, HCFSSNet is not latency-optimal. Its additional computational cost mainly comes from the omni-directional scan design, the DCT/IDCT-based frequency modulation, and the frequency-aware refinement introduced in the hyperprior pathway. Therefore, the current model should be viewed as a compact and competitive design point rather than a solution optimized for minimum runtime.

In future work, we will investigate lighter variants of HCFSSNet for lower-latency settings, including more efficient scan schedules, approximate frequency modulation schemes, and more efficient side-information refinement in the hyperprior pathway. We will also explore how the proposed design can be extended to video compression and related compression tasks.

\printcredits

\bibliographystyle{cas-model2-names}

\bibliography{reference}



\end{document}